\documentclass[journal]{IEEEtran}
\ifCLASSINFOpdf
\else
\fi
\usepackage{cite}
\usepackage{amsmath,amssymb,amsfonts}
\usepackage{algorithmic}
\usepackage{graphicx}
\usepackage{multirow}
\usepackage{textcomp}
\usepackage{hyperref}
\usepackage{filecontents}
\usepackage{lastpage}%
\usepackage{supertabular}%
\usepackage{tabu}%
\usepackage{multirow}%
\usepackage{diagbox}
\usepackage{longtable}%
\usepackage{color, colortbl}
\definecolor{Gray}{gray}{0.9}
\usepackage[algo2e,linesnumbered,ruled]{algorithm2e} 
\usepackage[numbers]{natbib}
\DeclareMathOperator*{\argmax}{argmax} 
\DeclareMathOperator*{\argmin}{argmin} 
\usepackage[flushleft]{threeparttable}
\begin{document}
\title{Learning and Sharing: A Multitask Genetic Programming Approach to Image Feature Learning}
	
\author{Ying Bi,~\IEEEmembership{Member,~IEEE}, Bing Xue,~\IEEEmembership{Member,~IEEE}, and Mengjie Zhang,~\IEEEmembership{Fellow,~IEEE,}
\thanks{The authours are with School of Engineering and Computer Science, Victoria University of Wellington, Wellington 6140, New Zealand (e-mail: ying.bi@ecs.vuw.ac.nz; bing.xue@ecs.vuw.ac.nz; mengjie.zhang@ecs.vuw.ac.nz).}
\thanks{Color versions of one or more of the figures in this paper are available online at http://ieeexplore.ieee.org.}}
\markboth{IEEE transactions on xxxx,~Vol.~XX, No.~X, Month~year}
{Bi \MakeLowercase{\textit{et al.}}: paper title}
\maketitle
\begin{abstract}
Using evolutionary computation algorithms to solve multiple tasks with knowledge sharing is a promising approach. Image feature learning can be considered as a multitask problem because different tasks may have a similar feature space. Genetic programming (GP) has been successfully applied to image feature learning for classification. However, most of the existing GP methods solve one task, independently, using sufficient training data. No multitask GP method has been developed for image feature learning. Therefore, this paper develops a multitask GP approach to image feature learning for classification with limited training data. Owing to the flexible representation of GP, a new knowledge sharing mechanism based on a new individual representation is developed to allow GP to automatically learn what to share across two tasks and to improve its learning performance. The shared knowledge is encoded as a common tree, which can represent the common/general features of two tasks. With the new individual representation, each task is solved using the features extracted from a common tree and a task-specific tree representing task-specific features. To learn the best common and task-specific trees, a new evolutionary process and new fitness functions are developed. The performance of the proposed approach is examined on six multitask problems of 12 image classification datasets with limited training data and compared with three GP and 14 non-GP-based competitive methods. Experimental results show that the new approach outperforms these compared methods in almost all the comparisons. Further analysis reveals that the new approach learns simple yet effective common trees with high effectiveness and transferability. 
\end{abstract}

\begin{IEEEkeywords}Evolutionary Multitask; Knowledge Sharing; Genetic Programming; Feature Learning; Image Classification, Evolutionary Computation
\end{IEEEkeywords}
\IEEEpeerreviewmaketitle

\section{Introduction}\label{sec:introduction}
Evolutionary multitask has become an increasingly popular topic in evolutionary computation (EC) and has been successfully applied to solve many problems such as function optimisation \cite{gupta2015multifactorial,feng2018evolutionary,ding2017generalized}, combinatorial optimisation \cite{gong2019evolutionary,feng2019solving}, complex engineering design \cite{gupta2015multifactorial}, and multiobjective optimisation \cite{gupta2016multiobjective,lin2019multi}. Evolutionary multitask \cite{gupta2015multifactorial} aims to use EC techniques to simultaneously solve multiple tasks, where the knowledge learned from different tasks can be implicitly or explicitly shared/transferred during the evolutionary process. Compared with traditional evolutionary single-tasking methods, evolutionary multitask methods often have faster convergence speed to find better solutions \cite{gupta2015multifactorial,bali2019multifactorial}. However, 
the potential of evolutionary multitask has not been comprehensively investigated in image feature learning. 

Image feature learning is the task of automatically learning informative features from images to solve a task, such as image classification \cite{shao2014feature, ma2020mdfn}. It is an essential step to image analysis, which has a wide range of real-world applications in many fields, including medicine, robotics, remote sensing, and security \cite{ain2020generating, al2017automatically,kemker2018low, chen2016bridge}. However, image feature learning is challenging because of the high variations crossing images and a large search space. The task becomes even more difficult when only a small number of training instances are available to evaluate the performance of the learned features during the learning process. A small number of training instances often lead to poor generalisation performance \cite{kemker2018low}. 

Many image feature learning methods have been developed for image classification \cite{jing2020self}. Most of them are based on neural networks (NNs), particularly convolutional neural networks (CNNs) \cite{kemker2018low, jing2020self}. However, these methods have limitations, such as require a large number of training instances and have poor interpretability. Except for NN-based methods, EC-based methods have also been developed for image feature learning in classification. The most commonly used methods are genetic programming (GP)-based methods \cite{al2019survey}. Unlike NN-based methods, GP can use existing image-related operators as internal nodes to automatically evolve tree-like solutions with potentially good interpretability, such as in \cite{shao2014feature, bi2019tevc}. Therefore, this paper uses GP to achieve image feature learning.

Image feature learning can be considered as a multitask problem because similar or related tasks may have a similar or common feature space. For example, to classify different texture image datasets, texture features, such as local binary patterns (LBP), are effective features. It is possible to simultaneously learn LBP-like features for different texture image classification tasks. Existing work has empirically shown the effectiveness of multitask feature learning with different assumptions on the task relatedness, such as in \cite{argyriou2007multi,kang2011learning}. However, most of the existing work is based on sparse representation learning and is not for image data. This paper addresses the multitask image feature learning task by automatically and simultaneously finding multiple optimal sets of features for diff image classification tasks. There could be multiple alternative ways to it. However, this has never been investigated using EC techniques. 


Existing EC methods have shown their potential of simultaneously solving two tasks and achieved better learning performance and/or faster convergence speed than solving those tasks individually  \cite{gupta2015multifactorial,feng2018evolutionary,ding2017generalized}. In evolutionary multitask, to avoid negative knowledge transfer/sharing crossing multiple tasks is important to improve the learning performance. Furthermore, image feature learning has two phases, i.e., the training/learning phase and the test phase. The training/learning phase can be achieved via the evolutionary process, while the test phase is after the evolutionary process. The test phase examines the performance of the learned features or models on unseen data, where is the generalisation performance. The learned model may memorize all the instances used in training and achieve high training performance, but have poor generalisation performance, which is known as overfitting. The overfitting issue is a common and open issue in machine learning. This is quite different from most optimisation problems solved by most of existing evolutionary multitask methods. Therefore, it is necessary to develop a new approach with considerations of how to improve the generalisation performance when addressing multitask image feature learning. 

GP has been successfully applied to feature learning for image classification \cite{al2019survey}. GP uses a tree-based variable-length representation and can automatically evolve trees/models that learn effective features from images for classification. Existing work on GP-based feature learning has achieved promising classification performance and potential interpretability, such as in \cite{bi2019tevc, shao2014feature}. However, existing GP-based image feature learning methods independently solve each task. To the best of our knowledge, there has not been any multitask GP method developed for image feature learning. Furthermore, most of existing GP-based methods evaluate the learned features using \emph{sufficient} training instances. However, training data is not always sufficient such as in the medicine and security domains, and/or need large manual effort to label. It is necessary to investigate whether GP-based methods can learn effective features using limited training data.

The overall goal of this paper is to further explore the capability of GP by developing a multitask GP approach to feature learning for image classification with limited training data. The new approach is termed as KSMTGP, indicating MultiTask GP with a new Knowledge Sharing mechanism. To avoid negative knowledge transfer/sharing and make the best use of the flexible representation of GP, a new explicit knowledge sharing mechanism and a new individual representation will be developed the KSMTGP approach. The main idea is to use KSMTGP to learn a common representation crossing two tasks and two task-specific representations in the form of trees. Each task will be solved by using two GP trees, i.e., a common tree and a task-specific tree. To achieve this, a new evolutionary search process and a new fitness function will be developed in KSMTGP to search for the best common and task-specific trees with variable lengths. The performance of the proposed KSMTGP approach will be evaluated on 12 image classification datasets with limited training data, i.e., six multitask feature learning problems. The performance of KSMTGP will be compared with a number of competitive methods to examine its effectiveness. 

The contributions of this paper are summarised as follows.
\begin{enumerate}
	\item The new KSMTGP approach is developed to achieve multitask feature learning for image classification. KSMTGP can achieve \emph{explicit} knowledge sharing in a form of GP tree with a variable length crossing two tasks. It is noted that most of existing evolutionary multitask methods cannot directly achieve explicit knowledge sharing. The new approach is able to  automatically determine what to share via learning and improve its performance by sharing. More importantly, the shared knowledge, i.e., the common tree, can be variable lengths and shapes as GP has a flexible variable-length representation.

	
	\item A multi-tree representation is proposed in KSMTGP to achieve explicit knowledge sharing. The multi-tree representation includes three trees, i.e., a common tree and two task-specific trees. Each task is solved by using two trees, i.e., a common tree and its corresponding task-specific tree. The common tree represents the common knowledge crossing two tasks, while the task-specific trees represent the knowledge specifically for solving the single task. 
	
	\item A new evolutionary search process is developed to search for the best common tree and the best task-specific trees for the two tasks. Similar to co-evolution, the new GP method searches for the best common tree and selects the best common tree coupled with a task-specific tree. This design narrows the search space and can find the best individuals of two trees for every single task effectively. 
	
	\item A new fitness function based on the classification performance of two tasks and the tree size is developed to evaluate the performance of the common tree. This new fitness function allows the new approach to learning simple yet effective common trees crossing two tasks. A fitness function is defined to effectively evaluate the fitness of GP individuals to find the best individuals of two trees for every single task. 
	
	\item The proposed KSMTGP approach is able to achieve better generalisation performance than two single-tasking GP methods, the multifactorial GP method, and 14 non-GP-based competitive methods on 12 image classification datasets with limited training data. 
\end{enumerate}
\section{Background and Related Work}

\subsection{Multitask Image Feature Learning}
Evolutionary multitask \cite{gupta2015multifactorial} is to simultaneously solve multiple problems using EC techniques with knowledge sharing during the evolutionary process. An evolutionary multitask optimisation problem has multiple tasks, denoted as $\{T_1,\ T_2,\ \dots,\ T_K\}$, where $K$ indicates the number of total tasks and $T_k$ indicates the $k$th task. Each task $T_k$ has a search space of $X_k$ and an objective function $f_k$. A multitask problem is to find the optimal solution $x^*_k$ for each task, concurrently, which is denoted as
\begin{equation}
\{x^*_1, \dots, x^*_K\} = \{\argmin f_1(x_1), \dots, \argmin f_K(x_K) \}.
\end{equation}

This paper proposes an evolutionary multitask approach to feature learning for image classification. An image classification task often has a training set $\mathcal{D}_{train}$ and a test set $\mathcal{D}_{test}$. The training set has a number of labelled images, i.e., $\mathcal{D}_{train}= \{(x_0, y_0), \dots, (x_n, y_n)\}$. The test set has a number of unlabelled images, i.e., $\mathcal{D}_{test}= \{z_0, \dots, z_m\}$. $x\in \mathbb{R}^{w\times h\times g}$ and $z\in \mathbb{R}^{w\times h\times g}$ denote a grey scale $(g=1)$ or colour $(g=3)$ image with a size of $w\times h$. $y \in \mathbb{R}^C$ denotes the class label. An image feature learning task is to find the optimal feature set transformed via a function/model $\Phi$ that can maximise a performance measure $\mathcal{L}$ on $\mathcal{D}_{train}$ as
\begin{equation}
\Phi^*= \argmax\mathcal{L}(\Phi, \mathcal{D}_{train}).
\end{equation}

In this study, a multitask problem of image feature learning for classification is to simultaneously find multiple optimal feature sets transformed by different functions/models with the goal of maximising a performance measure, which can be denoted as
\begin{multline}
\{\Phi^*_1, \dots, \Phi^*_K\}=\\
 \{\argmax\mathcal{L}_1(\Phi_1, \mathcal{D}^1_{train}), \dots, \argmax\mathcal{L}_K(\Phi_K, \mathcal{D}^K_{train})\}
\end{multline}

Typically, a feature learning problem has two phases, training/learning and test phases. The test phase is after the learning process. When $\{\Phi^*_1, \dots, \Phi^*_K\}$ are obtained, they will be applied to classify images in the test sets, i.e., $\{\mathcal{D}^1_{test}, \dots, \mathcal{D}^K_{test}\}$, respectively, and the classification performance on the test set is used to show the generalisation performance. For an image classification task, various methods can be used to find the optimal $\Phi$. This paper uses GP as the main method because of its flexible representation, which will be introduced in the following subsection. This paper aims to further explore the capability of GP for feature learning in image classification. 

\subsection{Genetic Programming}
\begin{figure}[htbp]
	\centering
	\vspace{-4mm}
	\includegraphics[width=0.8\linewidth]{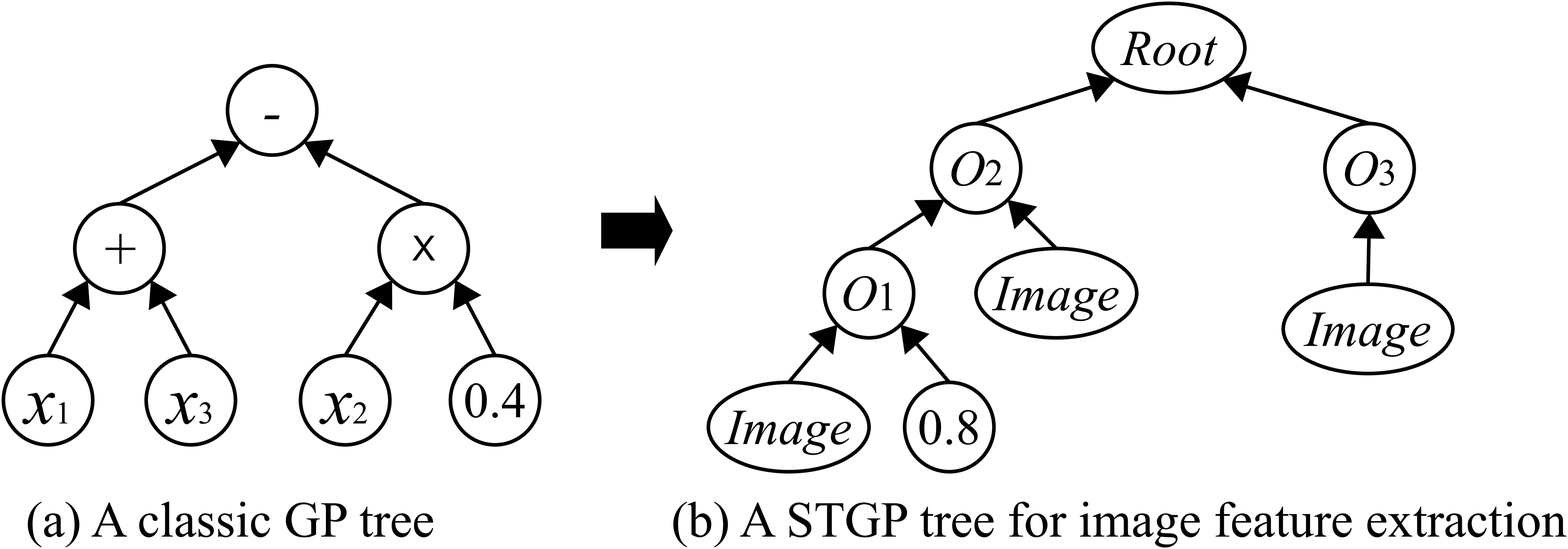}
	\vspace{-4mm}
	\caption{Example GP trees.}
	\label{fig:exampletree2}
	\vspace{-6mm}
\end{figure}

GP is an EC algorithm that is able to automatically evolve computer programs to solve a problem. Typically, GP uses a tree-based representation with variable lengths, which is different from the other EC methods such as genetic algorithms (GAs) and particle swarm optimisation (PSO) with a vector/string-based presentation typically with a fixed length. An example GP tree is shown in Fig. \ref{fig:exampletree2}(a). In this tree, the functions $+$, $-$ and $\times$ form the internal nodes and the features/variables/constants $x_1$, $x_2$, $x_3$, and $0.4$ form the leaf nodes. This example tree can be formulated as $(x_1+x_3)-(x_2\times 0.4)$. This tree representation is often used to solve regression, classification, feature construction, scheduling, and planning tasks \cite{al2019survey}. For some other tasks such as image analysis \cite{bi2018asurvey}, a different tree-based representation and some domain-specific operators are often used. Fig. \ref{fig:exampletree2}(b) shows an example GP tree with several specific operators $Root$, $O_1$, $O_2$ and $O_3$ and terminals $Image$ and $0.8$. This example tree can be formulated as $Root(O_2(O_1(Image,\ 0.8), Image),\ O_3(Image))$. This representation allows GP to use many domain-specific operators to evolve solutions with variable depths/lengths to solve different tasks. Examples of such GP trees to solve image classification and fault diagnosis can be found in \cite{bi2019tevc,bo2020Automatic}
 
\subsection{Related Work}
\subsubsection{GP for Image Feature Learning} 
\citeauthor{Atkins} \cite{Atkins} developed a multi-tier GP approach to achieve image filtering, region detection, feature extraction, and image classification, simultaneously. This methods has achieved better classification performance than the methods using raw pixels or statistics-based features on several datasets. \citeauthor{al2017automatically} \cite{al2017automatically} proposed a GP-based method to evolve texture descriptors for texture image classification. This method used a specific tree representation to detect features from keypoints and employed a similar way to LBP to generate a feature vector for classification. The results showed that this method achieved better performance than the other methods using different manually extracted texture features. In \cite{al2017keypoints}, an improved version of the method in \cite{al2017automatically} was proposed to allow GP to evolve a dynamic number of features for texture classification. This method had several specific root nodes to produce a flexible number of features. This method has achieved better classification performance than the method in \cite{al2017automatically} and the other methods using different texture features. In \cite{bi2019tevc}, a FGP method with a flexible program structure was proposed to automatically learn various types and numbers of features for image classification. The FGP method used many image-related operators to build the internal nodes of GP trees and learned domain-specific features. The FGP method has achieved better classification performance than many image classification methods, including traditional methods using manually extracted features and neural network-based methods, on different image classification tasks such as texture classification, scene classification and object classification. \citeauthor{ain2020generating} \cite{ain2020generating} proposed a GP method with a multi-tree representation to automatically learn effective features from various types of preextracted features for skin cancer image classification. This method has achieved significantly better performance than several GP methods and six most commonly used classification algorithms on two skin image datasets. These aforementioned GP methods focus on how effective image features are learned and generated by developing various GP representations for classification. Most of these methods tackled with a limited type of tasks and none of them has been applied to simultaneously solve two image classification tasks.

\subsubsection{Evolutionary Multitask}
Evolutionary multitask has become an increasingly popular research topic in EC and has been widely applied to solve many tasks. \citeauthor{gupta2015multifactorial} \cite{gupta2015multifactorial} proposed the early paradigm of evolutionary multitask in the field of EC, i.e., the multifactorial evolutionary algorithm (MFEA) to solve multiple problems, simultaneously. The main idea of MFEA is to use a unified search space to represent the population, which is split into different groups based on skill factors, where each group deals with a task. The whole population is evolved during the evolutionary process, in which the genetic materials are implicitly transferred via assortative mating and vertical cultural transmission. The experimental results showed that MFEA improved the convergence speed by simultaneously optimising two tasks. In \cite{gupta2016multiobjective}, the potential ability of MEFA was further explored by proposing a MO-MFEA method to simultaneously solve multiple multi-objective optimisation problems. The effectiveness of automatic implicit transfer has been shown on several multi-objective optimisation problems and real-world applications. \citeauthor{feng2018evolutionary} \cite{feng2018evolutionary} proposed an evolutionary multitask framework, i.e., EMT, with explicit knowledge sharing, which was achieved by using a denoising autoencoder. Unlike MFEA, which uses a unified search space for solving different tasks simultaneously, the EMT method uses independent individual representation for each task. The experimental results on single and multi-objective multitask optimisation benchmarks showed that EMT was more effective than the single-tasking method and MEFA by using explicit knowledge sharing. 

\citeauthor{zhou2020toward} \cite{zhou2020toward} developed an adaptive knowledge transfer strategy in MEFA, i.e., MEFA-AKT, for single- and multi-objective optimisation. The adaptive knowledge transfer is achieved by changing the crossover operators in MEFA using the information collected from the evolutionary process. The experimental results showed that MEFA-AKT achieved better performance than MEFA. In \cite{gong2019evolutionary}, the computing resources allocation problem in MEFA was investigated and a new strategy was proposed to dynamically allocate the resources to the tasks according to their complexities. The results show the effectiveness of the proposed method in comparisons with MEFA. \citeauthor{zheng2019self} \cite{zheng2019self} proposed a self-regulated evolutionary multitask optimisation framework, i.e., SREMTO with the consideration of tasks relatedness. Instead of factorial rank in MEFA, this method uses an ability vector to denote the ability of an individual handling each task. The superiority of SREMTO has been verified on two EMTO test suites.

\emph{Summary:}
The majority of existing evolutionary methods for multitask optimisation are developed for optimisation problems. The potential ability of these methods for learning problems, such as feature learning for image classification, has not been investigated. On the other hand, GP has been widely applied to feature learning for image classification. Existing GP methods learn features from sufficient training data. Furthermore, no multitask GP-based methods have been developed for feature learning for image classification. Therefore, this paper will fill this research gap by developing a multitask GP approach to feature learning for image classification tasks with limited training data. 

\section{The Proposed Approach}
The section describes the proposed GP approach (KSMTGP) to multitask feature learning for image classification with limited training data. It describes the main idea of knowledge sharing and how this is achieved in the feature learning process by developing a new individual representation. Then the evolutionary process, the fitness functions and the tree representation (the baseline method) are presented. 

\subsection{A New Individual Representation for Knowledge Sharing}



\begin{figure}
	\centering
	\includegraphics[width=\linewidth]{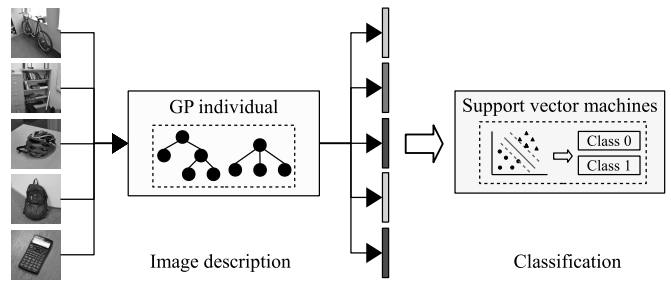}
	\vspace{-6mm}
	\caption{A general process of GP-based feature learning for image classification. The GP tree is able to transform images into a set of features, which are fed into a classification algorithm such as SVM for classification.}
	\label{fig:gpimageclassification}
	\vspace{-2mm}
\end{figure}

When applying GP to learn features, a tree often consists of many image-related operators and can be used to extract a set of features to describe an image, such as in \cite{shao2014feature, bi2019tevc}. In other words, each tree represents a possible image representation to solve a classification task. Given the GP trees, a general image classification process is shown in Fig. \ref{fig:gpimageclassification}, where a classification algorithm, i.e., support vector machine (SVM), is used to perform classification \cite{shao2014feature, bi2019tevc}. 

\begin{figure}
	\centering
	\vspace{-2mm}
	\includegraphics[width=0.9\linewidth]{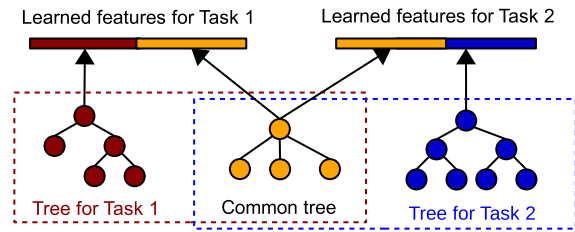}
	\vspace{-2mm}
	\caption{Individual representation of KSMTGP.}
	\label{fig:representation}
	\vspace{-6mm}
\end{figure}

The KSMTGP approach is developed to simultaneously search for the best individuals for the two feature learning tasks. To achieve this, a new individual representation is developed with the idea of knowledge sharing. The new representation is shown in Fig. \ref{fig:representation}, which is a multi-tree representation based on strongly typed GP \cite{montana1995strongly,poli08fieldguide}. The new representation of each individual in GP is encoded by three trees, i.e., the left (red) tree indicates a task-specific tree for task 1, the middle (yellow) tree indicates the common tree shared by the two tasks, and the right (blue) tree indicates a task-specific tree for task 2. In other words, the feature representation of the images for each task is the combination of common cross-task representation from the common tree and task-specific representation from the task-specific tree. The task-specific trees and the common trees are automatically learned through the evolutionary learning process with the use of different fitness functions.

Such a design is derived from the concept/assumption that two similar or related image classification tasks may have/share common feature representation. This concept/assumption can be seen from many other methods for multitask feature learning \cite{argyriou2007multi, li2017better} or the applications of pretrained CNN models as feature extractors to solve other image classification tasks \cite{shin2016deep}. In the proposed KSMTGP approach, the common tree is employed to describe common/general feature representation across two tasks. The common feature representation often improves the generalisation performance, particularly when the training data are limited \cite{argyriou2007multi}. The common tree, having a variable-length and shape, can be automatically evolved through the evolutionary process with the use of fitness evaluation and genetic operators. Two important characteristics are that the number of features described by the common tree varies with the tasks and the common tree can produce features from the given images of different sizes. These two characteristics enable the high adaptability of KSMTGP on various image classification tasks. 


Since the common tree represents the knowledge learned from the two tasks, which could be more general but less discriminative. To improve the discriminability of the features on each task, a task-specific tree is employed in KSMTGP to represent task-specific features. 
Two similar or related tasks may have not only common feature representation but also task-specific feature representation \cite{maurer2016benefit}. The combination of common and task-specific feature representations is expected to improve the learning performance on each task. Therefore, each task is solved by using two trees, i.e., a common tree and a task-specific tree. The features extracted by these two trees are combined for classification, as shown in Fig. \ref{fig:representation}. 

One important point is the search space of the evolutionary methods for solving two tasks \cite{gupta2015multifactorial}. The KSMTGP approach is based on GP. Typically, the search space of GP is related to the program structure, the function and terminal sets. Once these components are defined/set, they are often kept the same when applying GP to deal with the different image classification tasks, such as in \cite{shao2014feature, bi2019tevc, bi2020effective}. Therefore, there is no need to define a unified search space to use those GP methods to achieve multitask feature learning with the new individual representation. In the case that the search space of the two tasks is different, it is also easy to find the common settings of the function and terminal sets of the two tasks and use these settings in the search for the common tree. 



\subsection{Evolutionary Process}
To effectively search for the best common and task-specific trees, KSMTGP uses a new evolutionary search process similar to that of coevolution by using multiple populations. Coevolutionary algorithms have been widely used and achieved promising performance since they have more interactions between individuals \cite{antonio2017coevolutionary}. KSMTGP uses a population to search for the common tree and selects the best common tree to be combined with the task-specific trees at each generation. Based on the best common tree, two 2-tree populations are employed to search for the best trees for each task. These processes are conducted at each generation so that more possible combinations of these trees can be obtained and the performance on the tasks can be gradually and collaboratively improved. The overall process of KSMTGP is shown in Algorithm \ref{frameworkGP}. The inputs of the KSMTGP system are the training sets of two image classification tasks, i.e., $\mathcal{D}_{train}^1$ and $\mathcal{D}_{train}^2$.


The overall process of KSMTGP starts with randomly initialising three population, one common population and two task-specific populations for the two tasks. The common population aims to learn common knowledge from these two tasks, while the task-specific population aims to learn task-specific knowledge from each of these two tasks. This common population will be evaluated using a new fitness function, which is based on the training sets of the two tasks and a penalty of the tree size. Then the best tree of the common population is selected as the common knowledge learned from these two tasks. The best tree is combined with these two task-specific populations to form two 2-tree populations. Each individual of these two 2-tree populations has one tree from the task-specific population and one tree (i.e., the best tree) from the common population. The two 2-tree populations will be evaluated using the fitness function based on the training sets of these two tasks, respectively. 

The evolutionary process of these three populations (one common population and two task-specific population corresponding to the two tasks) is the same as that of the traditional GP method. The common population uses selection, subtree crossover and subtree mutation operators to generate a new population. These two task-specific populations use elitism, selection, subtree crossover and subtree mutation operators to generate new populations. Note that the common population generation does not have the elitism operator because only the best tree is selected to be combined with the task-specific population. So it is unnecessary to have an elitism operator. At each generation, the best individual (with two trees) for each task is updated. After the evolutionary process, the best individual of two trees for each task is returned.

\begin{algorithm2e}[htbp]
	\footnotesize
	\caption{KSMTGP}
	\label{frameworkGP}
	\SetKwData{Left}{left}\SetKwData{This}{this}\SetKwData{Up}{up}
	\SetKwFunction{Union}{Union}\SetKwFunction{FindCompress}{FindCompress}
	\SetKwInOut{Input}{Input }\SetKwInOut{Output}{Output }
	\Input{$\mathcal{D}_{train}^1$: the training set of task 1;\\ $\mathcal{D}^2_{train}$: the training set of task 2.}
	\Output{$Best\_Ind_1$: the best individual for task 1;\\ $Best\_Ind_2$: the best individual for task 2.}
	\BlankLine
	$P_0^{com}\leftarrow$ initialise a common population for the two tasks using the ramped half-and-half method;\\
	$P_0^{1}, P_0^{2} \leftarrow$ initialise two task-specific populations for the two tasks using the ramped half-and-half method;\\
	$g\leftarrow 0$;\\
	\While{$g\leq G$}{
		Evaluate the common population $P_g^{com}$ according to \textbf{Algorithm} \ref{FitnessEvaluationCommon};\\
		$ct_{best}\leftarrow$ select the best common tree from $P_g^{com}$;\\
		$(P_g^{1},\ ct_{best}), (P_g^{2},\ ct_{best}) \leftarrow$ form two 2-tree populations for the two tasks;\\
		Evaluate $(P_g^{1},\ ct_{best})$ and $(P_g^{2},\ ct_{best})$ on $\mathcal{D}_{train}^1$ and $\mathcal{D}_{train}^2$, respectively, according to \textbf{Algorithm} \ref{FitnessEvaluation};\\
		Update $Best\_Ind_1$ and $Best\_Ind_2$;\\
		\tcp{Generate a new common population using genetic operators}
		$P_{g+1}^{com} \leftarrow$ offspring generated from $P_{g}^{com} $ using selection, subtree crossover and subtree mutation;\\
		\tcp{Generate two new task-specific populations using genetic operators}	
		$P_{g+1}^{1} \leftarrow$ offspring generated from $P_g^{1}$ using elitism, selection, subtree crossover and subtree mutation;\\
		$P_{g+1}^{2} \leftarrow$ offspring generated from $P_g^{2}$ using elitism, selection, subtree crossover and subtree mutation;\\
		$(P_{g+1}^{1},\ ct_{best}), (P_{g+1}^{2},\ ct_{best}) \leftarrow$ form two multi-tree populations for the two tasks;\\
		$g\leftarrow g+1$;\\
	}
	Return $Best\_Ind_1$ and $Best\_Ind_2$.
\end{algorithm2e}

\vspace{-6mm}
\subsection{Fitness Evaluation}

\subsubsection{Fitness Evaluation for Common Population}
The common population aims to learn common and general knowledge crossing two tasks. A better common tree should not only achieve good performance on two tasks but also have less complexity. Many research works have shown that a simple model often has better generalisation performance than a complex model \cite{chen2017feature}. However, measuring the complexity of a GP tree with many image-related operators as internal nodes is not easy. For simplification, the tree size is used as an indicator of the tree complexity. With the integration of the classification performance on the two tasks and the tree size, a new fitness function is proposed to evaluate the performance of the common tree. The new fitness function to be maximised is defined in Eq. (\ref{com_fit}). The overall fitness evaluation process is described in Algorithm \ref{FitnessEvaluationCommon}.
\begin{equation}\label{com_fit}
f_{com} 
=\frac{1}{2}(acc_1+acc_2)-size
\end{equation}
\begin{equation}\label{acc}
acc =\frac{TP+TN}{TP+TN+FP+FN}*100
\end{equation}
where $K$ indicates the number of folds in $K$-fold cross validation. $acc_1$ and $acc_2$ indicate the average classification accuracy (\%) obtained using $K$-fold cross validation and a classification algorithm on the training sets of task 1 and task 2, respectively. $size$ denotes the tree size. $TP$, $TN$, $FP$, and $FN$ indicate true positive, true negative, false positive, and false negative, respectively. 

\begin{algorithm2e}[htbp]
	\footnotesize
	\caption{Fitness Evaluation for Common Population}
	\label{FitnessEvaluationCommon}
	\SetKwData{Left}{left}\SetKwData{This}{this}\SetKwData{Up}{up}
	\SetKwFunction{Union}{Union}\SetKwFunction{FindCompress}{FindCompress}
	\SetKwInOut{Input}{Input }\SetKwInOut{Output}{Output }
	\Input{$\mathcal{D}_{train}^1$: the training set of task 1; $\mathcal{D}_{train}^2$: the training set of task 2; $P^{com}$: the population to be evaluated.}
	\Output{$P^{com}$: the evaluated population.}
	\BlankLine
	\For{each tree $p$ in $P^{com}$}{
		\For{each task $t$ in $\{1, 2\}$}{
		\For{each image $i$ in $\mathcal{D}_{train}^t$}{
			$f_i=\{f_{i, 0},\ \dots,\ f_{i, m_p}\} \leftarrow$ features extracted from image $i$ using the tree of $p$;\\
		}
	$\mathcal{D}^{t}_{tr}\leftarrow$ the transformed training set with the features;\\
	$\mathcal{D}^{t}_{nrom}\leftarrow$ perform the min-max normalisation;\\
	$acc_t \leftarrow$ use SVM with $K$-fold cross-validation on the normalised training set $\mathcal{D}^{t}_{nrom}$ to obtain the average accuracy;\\
	}
$size \leftarrow len(p)$;\\
$f_{com}(p) \leftarrow \frac{1}{2}\sum_{t=1}^{2}acc_t-size$.
	}
\end{algorithm2e}

\subsubsection{Fitness Evaluation for Each Task} With the individual representation in KSMTGP, each task is solved by using two trees, one from the task-specific population and the other is the best common tree. In the fitness evaluation, the performance of the two trees is evaluated using a fitness function, which is defined in Equations (\ref{acc_tt1}) and (\ref{acc}). It calculates the classification accuracy according to the training set of each task. 
\begin{equation}\label{acc_tt1}
f_{t}=acc
\end{equation}

The overall fitness evaluation process is described in Algorithm \ref{FitnessEvaluation}. This process starts by using the best common tree ($ct_{best}$) to transform the whole training set into features. When evaluating each individual, the transformed features by $ct_{best}$ will be directly used rather than performing transformation again, which saves computational cost. 

\begin{algorithm2e}[htbp]
	\footnotesize
	\caption{Fitness Evaluation for Each Task}
	\label{FitnessEvaluation}
	\SetKwData{Left}{left}\SetKwData{This}{this}\SetKwData{Up}{up}
	\SetKwFunction{Union}{Union}\SetKwFunction{FindCompress}{FindCompress}
	\SetKwInOut{Input}{Input }\SetKwInOut{Output}{Output }
	\Input{$\mathcal{D}_{train}$: the training set; $(P,\ ct_{best})$: the population to be evaluated.}
	\Output{$(P,\ ct_{best})$: the evaluated population.}
	\BlankLine
	
\For{each image $i$ in $\mathcal{D}_{train}$}{
	$f_i=\{f_{i, 0},\ \dots,\ f_{i, m}\} \leftarrow$ features extracted from image $i$ using the tree of $ct_{best}$;\\
}

\For{each tree $p$ in $P$}{
\For{each image $i$ in $\mathcal{D}_{train}$}{
	$v_i=\{v_{i, 0},\ \dots,\ v_{i, n_p}\} \leftarrow$ features extracted from image $i$ using the tree of $p$;\\
	$\{v_i,\ f_i\}\leftarrow$ concatenating the features extracted by $p$ and $ct_{best}$ to form the feature set for describing image $i$;\\
}
$acc \leftarrow$ normalise the transformed training set and use SVM with $K$-fold cross-validation to obtain the average accuracy;\\
$f(p) \leftarrow acc$.
}
\end{algorithm2e}

In the fitness evaluation for common population and the population for each task, a linear SVM is employed to obtain the classification accuracy based on the features learned by GP. The linear SVM is employed because it is commonly used in GP-based methods for image classification \cite{bi2019tevc, shao2014feature}. The classification accuracy is obtained by performing $K$-fold cross validation on the training set. In this process, the whole training set is split into $K$ folds. The $k$th fold is used as an evaluation test set and the remaining $K-1$ folds are used as the evaluation training set. The evaluation training set is used to train SVM classifiers and the ``test" accuracy is obtained on the evaluation test set. This process repeats $K$ times until each fold is used as the evaluation test set once. The average ``test" accuracy of the $K$ folds are used in the fitness evaluation.

\subsection{Tree Structure and Baseline Method}
The algorithm framework of KSMTGP can cooperate with different GP tree structures that are able to transform images into features to achieve multitask feature learning for image classification. To investigate the performance of KSMTGP, this study uses a recently proposed tree structure of a baseline method, i.e., FGP in \cite{bi2019tevc}, because it is a very flexible representation to allow GP to learn variable numbers and types of features for different image classification tasks. The KSMTGP uses the same tree structure as FGP but a different individual representation, fitness functions and evolutionary search process. This section will describe the tree structure employed in KSMTGP. 

\subsubsection{Tree Structure} 
\begin{figure}[htbp]
	\centering
	\includegraphics[width=\linewidth]{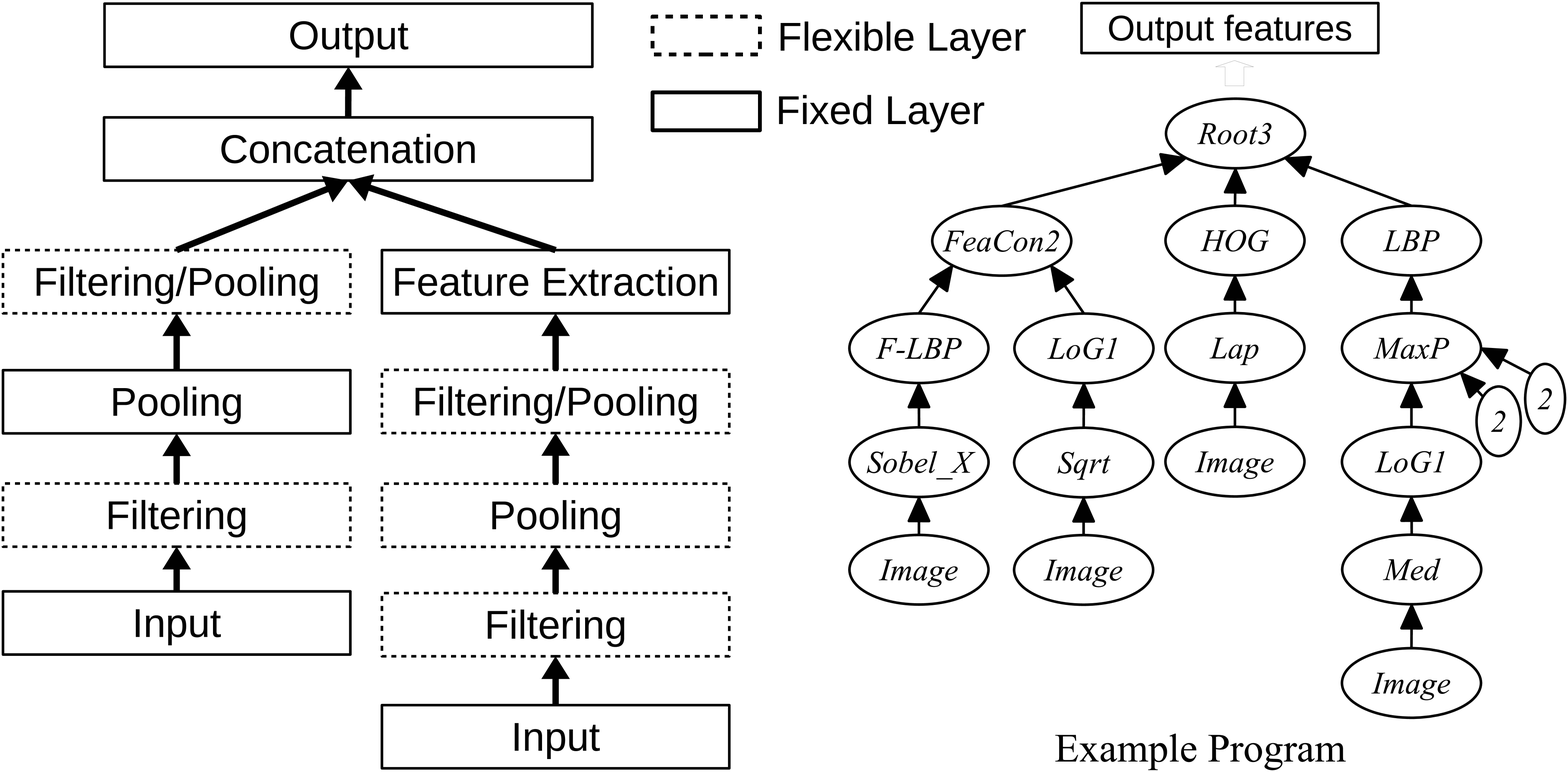}
	\vspace{-6mm}
	\caption{The tree structure and an example tree (adapted from \cite{bi2019tevc}).}
	\label{fig:fgpstructure}
	 	\vspace{-6mm}
\end{figure}

The tree structure and an example tree is shown in Fig. \ref{fig:fgpstructure}. The tree structure has a number of different layers, including an input layer, filtering layers, pooling layers, a feature extraction layer, a feature concatenation layer, and an output layer. Each layer has different functions for different purposes. This tree structure is able to produce variable-length solutions that can generate different types and numbers of features for describing images. 

\subsubsection{Function Set} The function set has filtering functions, a pooling function, feature extraction functions, and concatenation functions, which are listed in Table \ref{table:featurelearningfunc}. The filtering functions take images and parameters if required as inputs and output a new image being processed. The filtering functions include Gaussian filters ($Gau$), Gaussian derivatives ($GauD$), Gabor filters ($Gabor$), Laplacian filter ($Lap$), Laplacian of Gaussian filters with two different parameter values ($LoG1$ and $LoG2$), Sobel edge detectors ($Sobel$, $sobelX$, $SobelY$), median filter ($Med$), mean filter ($Mean$), min filter ($Min$), max filter ($Max$), LBP as a filter ($LBP${-}$F$), HOG as a filter ($HOG${-}$F$), weighted sum ($W${-}$Add$), weighted subtraction ($W${-}$Sub$), rectify linear unit ($ReLU$), and Sqrt ($Sqrt$). The pooling function is max-pooling with two parameters ($MaxP$). The feature extraction functions are commonly used feature extraction methods, which are $LBP$, $HOG$ and $SIFT$. The concatenation functions concatenate features or images into a feature vector. These functions are $FeaCon2$, $FeaCon3$, $Root2$, $Root3$, and $Root4$.

\begin{table}[htbp]
	\vspace{-2mm}
	\setlength{\tabcolsep}{2pt}
	\renewcommand{\arraystretch}{1.2}
	\footnotesize
	\vspace{-2mm}
	\caption{Function set}
	\vspace{-4mm}
	\begin{center}
		\begin{tabular}{p{0.235\linewidth}|p{0.73\linewidth}}
			\hline 
			\textbf{Layer}&\textbf{Function}\\
			\hline
			Filtering&$Gau$, $GauD$, $Gabor$, $Lap$, $LoG1$, $LoG2$, $Sobel$, $SobelX$, $SobelY$, $Med$, $Mean$, $Min$, $Max$, $LBP\text{-}F$, $HOG\text{-}F$, $W\text{-}Add$, $W\text{-}Sub$, $ReLU$, $Sqrt$\\
			\hline
			Pooling&$MaxP$\\
			\hline
			Feature Extraction&$SIFT, HOG, LBP$\\
			\hline
			Concatenation&$FeaCon2$, $FeaCon3$, $Root2$, $Root3$, $Root4$\\
			\hline		
		\end{tabular}	
		\label{table:featurelearningfunc}
	\end{center}
	\vspace{-4mm}
\end{table}

\subsubsection{Terminal Set}The terminal set has $Image$ indicating the input image, and the parameters (i.e., $\sigma$, $o_1$, $o_2$, $\theta$, $f$, $n_1$, $n_2$, $k_1$, and $k_2$) for particular filtering and pooling operators. The $\sigma$ terminal indicates the standard deviation of a Gaussian function in the $Gau$ and $GauD$ filters. It is in the range $[1,3]$. The $o_1$ and $o_2$ terminals denote the order of the Gaussian derivatives and are in the range $[0,2]$. The $\theta$ terminal represents the orientation of the $Gabor$ filter and is in the range $[0,~ 7\pi/8]$ with a step of $\pi/8$ \cite{liu2002gabor}. The $f$ terminal is the $Gabor$ filter and equals to $\frac{\pi}{2}/{\sqrt{2}^v}$, where $v$ is an integer in the range of $[0, 4]$ \cite{liu2002gabor}. The $n_1$ and $n_2$ terminals are parameters for the $W${-}$Add$ and $W${-}$Sub$ functions. They are in the range $[0,~1)$. The $k_1$ and $k_2$ terminals indicate the kernel size of the $MaxP$ function and are in the range $\{2, 4\}$.

\section{Experiment Design}

This section designs the experiments that are conducted to show the effectiveness of the proposed approach. It describes comparison methods, benchmark problems of image classification with limited training data, and parameter settings.

\subsection{Comparison Methods}
The performance of KSMTGP is compared with three GP-based methods, five classification algorithms, five traditional methods using manually designed features, and four methods based on CNNs, which are described as follows.

The three GP-based methods are a single-tasking GP method with a single tree representation (FGP proposed in \cite{bi2019tevc}), a single-task GP method with a multi-tree representation (MTFGP), and a multitask method (i.e., multifactorial GP, MFFGP). These three methods use the same tree representation as KSMTGP. MTFGP evolves two trees to solve a task and uses the genetic operators employed in \cite{ain2020generating} for a multi-tree representation. The MFFGP method is based on the multifactorial framework for solving multiple tasks \cite{gupta2015multifactorial}. MFFGP uses a different evolutionary process to search for the best solutions for the two tasks, simultaneously. Since no multitask GP methods proposed for image feature learning, the standard multitask framework is used in GP to investigate whether the proposed framework is better for image classification. 

	 
	

The five classification algorithms are SVM, random forest (RF) \cite{zhou2018deep}, k-nearest neighbour (KNN) \cite{al2017keypoints}, sparse representation-based classification (SRC) \cite{wright2008robust}, and linear discriminant analysis (LDA) \cite{belhumeur1997eigenfaces}. The aim is to investigate whether KSMTGP can learn features that are more effective than raw pixels for classification using limited training data. 

The five traditional methods using manually designed features are domain-independent features (DIF) \cite{zhang2003domain}, histogram features (Histogram) \cite{hassaballah2016image}, histogram of oriented gradients (HOG) \cite{dalal2005histograms}, local binary patterns (LBP) \cite{ojala2002multiresolution}, and scale-invariant feature transform (SIFT) \cite{lowe2004distinctive}. These methods use the corresponding features as inputs of a linear SVM to perform classification. They represent traditional image classification methods and the aim of comparisons is to show whether KSMTGP can learn better features with limited training data.
	
Two CNNs and two deep CNNs with transfer learning are used for comparisons. The two CNNs are LeNet \cite{lecun1998gradient} and a five-layer CNN (CNN-5) \cite{shao2014feature}. We do not directly train deep CNNs (including state-of-the-art CNNs) for comparisons due to a small number of training instances are used. Instead, the features extracted from pre-trained state-of-the-art deep CNNs, i.e., InceptionV3 \cite{szegedy2016rethinking} and InceptionResNetV2 \cite{szegedy2017inception}, are used for comparisons. These two CNNs were trained on ImageNet \cite{deng2009imagenet} and the extracted features are fed into SVMs for classification. From our preliminary experiments, we found the features extracted from InceptionV3 and InceptionResNetV2 achieved better results than features from other famous pretrained deep CNN models, e.g., VGGNet, ResNet and DenseNet. Therefore, we only choose these two best methods for comparisons. 

\subsection{Benchmark Problems}
\begin{figure}
	\centering
	\includegraphics[width=\linewidth]{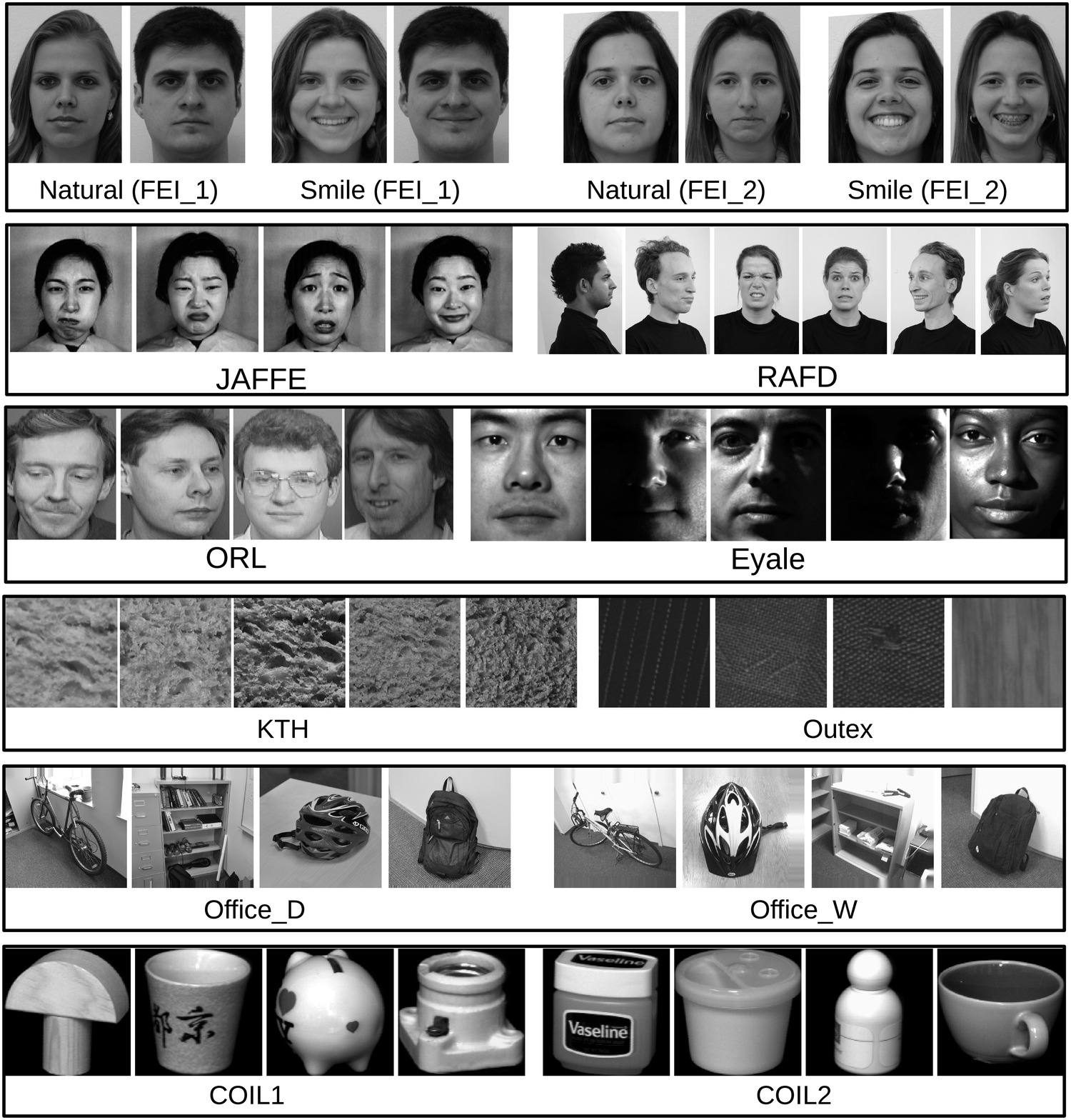}
	\vspace{-6mm}
	\caption{Example images from the 12 image classification datasets.}
	\label{fig:datasetsall}
	\vspace{-4mm}
\end{figure}

Due to no multitask image classification problems are available for examining the performance of the proposed method, we construct six problems according to the task type based on 12 different image classification datasets. 
The 12 datasets have various images sizes, numbers of classes and numbers of images. They are FEI\_1 \cite{thomaz2012fei}, FEI\_2 \cite{thomaz2012fei}, JAFFE \cite{lyons1998coding}, RAFD \cite{langner2010presentation}, ORL \cite{samaria1994parameterisation}, EYALE \cite{lee2005acquiring}, KTH \cite{mallikarjuna2006kth}, Outex \cite{ojala2002outex}, Office\_D \cite{saenko2010adapting}, Office\_W \cite{saenko2010adapting}, COIL1 \cite{nene1996columbia}, and COIL2 \cite{nene1996columbia}. These datasets are chosen because they represent typical image classification tasks, i.e., facial expression classification (FEI\_1, FEI\_2, JAFFE, and RAFD), face classification (ORL and EYALE), texture classification (KTH and Outex), and object classification (Office\_D, Office\_W, COIL\_1, and COIL\_2), and contain a wide range of image variations, including rotation, pose, illumination, scale, deformation, and occlusion variations. These factors will comprehensively show the effectiveness of the proposed approach. Example images from these datasets as shown in Fig. \ref{fig:datasetsall}. 

%
%
%
%

According to the types of tasks, six problems of multitask image feature learning are established to evaluate the performance of the proposed KSMTGP approach. The six problems are two facial expression classification problems, one face classification problem, one texture classification problem, and two object classification problems. The details of these six problems are listed in Table \ref{table:datasets_new}. As the proposed approach aims to learn effective features from limited training data, a small number of instances are randomly selected to form the training sets of these datasets. The fourth column of Table \ref{table:datasets_new} shows the number of training images in each class, which ranges from 5 to 25 except for the RAFD dataset. The RAFD dataset is a large dataset, having 1,000 images per class. Therefore, 10\% images are used as the training set and the remaining images are used as the test set as this study focuses on limited training data. Table \ref{table:datasets_new} also shows that the numbers of classes and the image sizes of the two tasks are different in most of the six multitask problems. To reduce the computational cost, some large images are resized to smaller images and the colour images are converted to gray-scale images.
 
\begin{table}[!]
\setlength{\tabcolsep}{0.6em} 
\caption{Summary of the datasets}
	\vspace{-4mm}
	\begin{center}
		\begin{tabular}{|p{0.14\linewidth}|p{0.155\linewidth}|p{0.07\linewidth}|p{0.08\linewidth}|p{0.23\linewidth}|p{0.08\linewidth}|}
			\hline 
			&Datasets&\#Class &Image Size&Train Set (Per Class)& Test Set\\
			\hline	
		\multirow{2}{*}{\textbf{Problem 1}} &FEI\_1&2&40$\times$60&50 (25 images)&150\\
		&FEI\_2 &2&40$\times$60&50 (25 images)&150\\ \hline
			\multirow{2}{*}{\textbf{Problem 2}} &JAFFE&7&64$\times$64&70 (10 images)&143\\
	&RAFD&8&40$\times$60&800 (100 images)&7,240\\ \hline
		\multirow{2}{*}{\textbf{Problem 3}} &ORL&40&60$\times$50
			&200 (5 images)&200\\
&EYALE&38&45$\times$50&380 (10 images)&2,044\\\hline
		\multirow{2}{*}{\textbf{Problem 4}} &KTH&10&50$\times$50&200 (20 images)&610\\
	&Outex&24&64$\times$64&480 (20 images)&3,840\\ \hline
	\multirow{2}{*}{\textbf{Problem 5}} &Office\_D&31&50$\times$50&155 (5 images)&343\\
&Office\_W&31&50$\times$50&155 (5 images)&640\\ \hline
	\multirow{2}{*}{\textbf{Problem 6}} &COIL\_1&10&32$\times$32&100 (10 images)&620\\
	&COIL\_2&10&32$\times$32&100 (10 images)&620\\ 
			\hline
		\end{tabular}	
		\label{table:datasets_new}
	\end{center}
	\vspace{-6mm}
\end{table}


\subsection{Parameter Settings}
 The parameters for KSMTGP are based on the commonly used settings of the GP community \cite{bi2020effective,iqbal2017cross}. In KSMTGP, the population size is 100, i.e, each population has a population size of 100. The maximum number of generation is 50. The crossover, mutation and elitism rates are 0.8, 0.19 and 0.01, respectively. The maximum tree depth is 8 and the minimal tree depth is 2. The tournament selection with a size of five is used for selecting individuals for crossover and mutation. For fair comparisons, the three GP-based methods (FGP, MTFGP and MFFGP) use the same parameter settings as KSMTGP. FGP and MTFGP deal with each task solely and use a population size of 100. MFFGP is a multitask method so that it uses a population size of 200 for the two tasks. In the fitness evaluation of these GP-based methods, SVM with $K$-fold cross validation is used to calculate the accuracy on the training set. We set $K=3$ for $K$-fold cross validation as the number of training instances is small.
 
 The parameter settings for the non-GP-based methods follow the commonly used settings \cite{young2018deep, zhou2018deep, al2017keypoints, chollet2015keras}. The key parameters of these methods are listed as follows. The number of neighbours in KNN is set to one \cite{al2017keypoints} and the penalty parameter C in SVM is set to one \cite{young2018deep}. In RF, the number of trees is 500 and the maximum tree depth is 100 \cite{zhou2018deep}. In LeNet and CNN, the activation function is ReLU and the classification method is softmax. In LeNet and CNN, the number of epochs is set to 100 and the batch size is set to 20 due to a small number of training instances employed.

\begin{table*}[!htbp]
\caption{Classification Accuracy (\%) of FGP, MTFGP, MFFGP, and KSMTGP on the twelve image classification datasets}
\vspace{-4mm}
\begin{center}
	\begin{tabular}{|l|l|ll|ll|ll|ll|}
	\hline 
\multirow{2}{*}{Problem}&\multirow{2}{*}{Dataset}&\multicolumn{2}{c|}{\textbf{FGP}}&\multicolumn{2}{c|}{\textbf{MTFGP}}&\multicolumn{2}{c|}{\textbf{MFFGP}}&\multicolumn{2}{c|}{\textbf{KSMTGP}}\\ 
&&Max&Mean $\pm$ St.dev&Max&Mean $\pm$ St.dev&Max&Mean $\pm$ St.dev&Max&Mean $\pm$ St.dev\\	\hline	
\textbf{Problem 1}&FEI\_1&89.33&84.56 $\pm$ 3.50 +&\textbf{94.00}&86.05 $\pm$ 4.27 =&91.33&86.20 $\pm$ 2.83 =&\textbf{94.00}&\textbf{87.91 $\pm$ 3.39}\\
&FEI\_2&92.00&87.16 $\pm$ 3.03 =&92.67&86.55 $\pm$ 2.80 =&92.67&87.98 $\pm$ 3.46 =&\textbf{94.67}&\textbf{88.24 $\pm$ 3.46}\\ 
\textbf{Problem 2} &JAFFE&66.43&61.65 $\pm$ 3.35 +&67.83&58.83 $\pm$ 4.46 +&67.13&60.58 $\pm$ 3.23 +&\textbf{68.53} &\textbf{64.13 $\pm$ 2.60}\\
&RAFD&49.83&44.82 $\pm$ 2.36 +&49.64&\textbf{46.95 $\pm$ 2.32} =&\textbf{50.26}&43.14 $\pm$ 3.82 +&49.82&46.45 $\pm$ 1.66 \\
\textbf{Problem 3}&ORL&100.0&\textbf{99.33 $\pm$ 0.45} =&100.0&99.27 $\pm$ 0.57 =&100.0 &99.17 $\pm$ 0.62 =&100.0&99.27 $\pm$ 0.53\\
&EYALE&99.71&98.52 $\pm$ 1.48 =&\textbf{99.80}&99.04 $\pm$ 0.50 +&99.71&98.76 $\pm$ 1.25 =&\textbf{99.80}&\textbf{99.35 $\pm$ 0.30}\\
\textbf{Problem 4}&KTH&94.59&92.04 $\pm$ 1.32 +&95.08&93.58 $\pm$ 0.96 =&\textbf{96.23}&93.05 $\pm$ 1.87 +&96.07&\textbf{94.14 $\pm$ 1.11}\\
&Outex&98.96&97.73 $\pm$ 0.74 +&99.40&98.70 $\pm$ 0.54 =&99.24&98.05 $\pm$ 0.72 +&\textbf{99.45}&\textbf{98.76 $\pm$ 0.30}\\
\textbf{Problem 5}&Office\_D&60.35&57.26 $\pm$ 1.87 +&60.64&57.85 $\pm$ 2.29 =&60.35 &57.47 $\pm$ 2.40 +&\textbf{63.27}&\textbf{59.16 $\pm$ 2.10}\\
&Office\_W&61.41&56.10 $\pm$ 2.12 +&61.09 &56.80 $\pm$ 2.00 =&60.31 &56.09 $\pm$ 1.88 +&\textbf{61.88}&\textbf{57.22 $\pm$ 1.74} \\
\textbf{Problem 6}&COIL\_1&93.71&91.52 $\pm$ 1.33 +&\textbf{94.68}&92.40 $\pm$ 1.33 =&\textbf{94.68}&92.10 $\pm$ 1.08 +&94.52&\textbf{92.89 $\pm$ 0.81}\\
&COIL\_2&100.0 &98.84 $\pm$ 0.92 +&100.0 &98.80 $\pm$ 1.25 +&100.0 &98.43 $\pm$ 1.69 +&100.0&\textbf{99.75 $\pm$ 0.41}\\ \hline 
\multicolumn{2}{|c|}{\textbf{Overall}}&&9+, 3=&&3+, 9=&&8+, 4=&&\\
	\hline
\end{tabular}	
\label{table:testresults}
\end{center}
	\vspace{-6mm}
\end{table*}

The GP-based methods are implemented using the DEAP (\emph{Distributed Evolutionary Algorithm in Python}) \cite{DEAP_JMLR2012} package. The classification algorithms in the GP methods and the non-GP-based methods are based on the \emph{scikit-learn} package \cite{scikit-learn}. The two CNN methods, the InceptionV3 method and the InceptionResNetV2 method are based on the \emph{Keras} package \cite{chollet2015keras}. To avoid experimental bias, the experiments of each algorithm on every dataset/task run independently 30 times using different random seeds. 

\section{Results and Discussions}

As the main problem is image classification, this section firstly analyses and discusses the classification (generalisation) performance of KSMTGP by comparing it with the GP-based and non-GP-based comparison methods. Additional to the classification performance, we also investigate the other aspects of KSMTGP, including training performance, convergence behaviour, computational time, and the number of learned features. This section also deeply analyses these aspects of KSMTGP by comparing it with the GP-based methods to provide more insights on its effectiveness. 

\subsection{Classification Performance on the Test Sets}

This section compares the classification performance of KSMTGP with that of the GP-based and non-GP-based comparison methods on the six multitask problems of image classification. The classification accuracy (\%) is used as the performance measure because it is the most commonly used. The results are listed in Tables \ref{table:testresults} and \ref{table:testresultsOther1}. To show the significance of classification performance improvement, Wilcoxon rank-sum test with a 0.05 level is used to compare the proposed KSMTGP method with each of the comparison methods. In these tables, ``+" denotes KSMTGP achieves significantly better performance and ``{--}" denotes KSMTGP achieves significantly worse performance than the compared method. ``=" means there is no significant difference. 
\subsubsection{Overall Performance} From Tables \ref{table:testresults}, it can be found that the proposed KSMTGP approach outperforms the single-tasking GP (FGP), multi-tree GP (MTGP) and multi-factorial GP (MFFGP) methods by achieving similar or significantly better classification performance on six multitask problems of 12 image classification datasets, i.e., significantly better in 20 comparisons and similar 16 in out of the total 36 comparisons. Table \ref{table:testresultsOther1} shows the proposed KSMTGP approach outperforms 14 non-GP-based methods in almost all the comparisons on the 12 image classification datasets as it achieves significantly better performance in 160 comparisons out of the total 168 comparisons. The results suggest that KSMTGP is an effective approach to automatically learning informative image features from two related or similar tasks for image classification. More detailed discussions and comparisons will be described in the following subsections.
 
\subsubsection{Comparisons with Single-Tasking GP (FGP)} The results in Table \ref{table:testresults} show that KSMTGP achieves significantly better performance than FGP on nine datasets and similar performance to it on three datasets. On these three datasets, the mean test accuracy (\%) obtained by KSMTGP is higher than that by FGP, although their performance has no significant difference. Comparing to solving each image classification task individually, jointly solving two similar or related tasks using the proposed KSMTGP approach is more effective. The multi-tree representation and the knowledge sharing mechanism in KSMTGP are the reasons for the improvement of generalisation performance, which will be further analysed and discussed in the following sections. 

\subsubsection{Comparisons with Single-Tasking GP with a Multi-Tree Representation (MTFGP)} Compared with MTFGP, the proposed KSMTGP approach achieves significantly better performance on three datasets and similar performance on six datasets. Specifically, KSMTGP achieves better mean accuracy than MTFGP on ten datasets, although their classification results of the 30 runs are not significantly different. Comparing with MTFGP and FGP, it can be found that a multi-tree representation is more effective than a single-tree representation in GP for feature learning to image classification in most cases. A multi-tree representation allows a GP individual to represent more information about the images or the class distributions. However, a multi-tree representation does not necessarily improve the classification performance of GP due to the issue of overfitting when the number of training instances is small. For example, MTFGP achieves worse results than FGP on the JAFFE, ORL and COIL\_2 datasets. The same as MTFGP, KSMTGP also uses an individual of two trees to representation a solution. Differently, KSMTGP learns a common tree from two similar or related tasks as a part of the multi-tree representation to represent more general knowledge/features, which can improve its generalisation performance. 

\begin{table*}[htbp]
	\caption{Classification Accuracy (\%) of KSMTGP and non-GP-based comparison methods on the twelve image classification datasets}
	\vspace{-4mm}
	\begin{center}
		\begin{tabular}{|l|l|l|l|l|l|l|}
			\hline 
&Mean $\pm$ St.dev&Mean $\pm$ St.dev&Mean $\pm$ St.dev&Mean $\pm$ St.dev&Mean $\pm$ St.dev&Mean $\pm$ St.dev\\\hline
\cellcolor[gray]{0.9}Methods&\cellcolor[gray]{0.9}\textbf{FEI\_1}&\cellcolor[gray]{0.9}\textbf{FEI\_2}&\cellcolor[gray]{0.9}\textbf{JAFFE}&\cellcolor[gray]{0.9}\textbf{RAFD}&\cellcolor[gray]{0.9}\textbf{ORL}&\cellcolor[gray]{0.9}\textbf{EYALE}\\ \hline				
SVM&78.02 $\pm$ 0.12 +& 85.35 $\pm$ 0.12 +& \textbf{64.94
$\pm$ 0.30} {--}& 32.91 $\pm$ 0.63+&97.00 $\pm$ 0.00 +&81.66 $\pm$ 0.09 +\\
RF&86.56 $\pm$ 1.33 +&82.47 $\pm$ 1.04 +&50.12 $\pm$ 1.12 +&24.61
$\pm$ 0.27 +&96.28 $\pm$ 0.63 +&74.35 $\pm$ 0.36 +\\
KNN&48.67 $\pm$ 0.00 +&48.00 $\pm$ 0.00 +&14.69 $\pm$ 0.00 +&13.36 $\pm$ 0.00 +&82.50 $\pm$ 0.00 +&27.74 $\pm$ 0.00 +\\
SRC&82.00 $\pm$ 0.00 +&84.00 $\pm$ 0.00 +&56.64 $\pm$ 0.00 +&29.82 $\pm$ 0.00 +&92.50 $\pm$ 0.00 +&90.66 $\pm$ 0.00 +\\
LDA&86.67 $\pm$ 0.00 +&85.33 $\pm$ 0.00 +&51.05 $\pm$ 0.00 +&31.05 $\pm$ 0.00 +&97.00 $\pm$ 0.00 +&81.56 $\pm$ 0.00 +\\
DIF&48.67 $\pm$ 0.00 +&54.00 $\pm$ 0.00 +&19.58 $\pm$ 0.00 +&18.44 $\pm$ 0.00 +&83.50 $\pm$ 0.00 +&19.23 $\pm$ 0.00 +\\
Histogram&53.33 $\pm$ 0.00 +&48.00 $\pm$ 0.00 +&18.88 $\pm$ 0.00 +&13.14 $\pm$ 0.00 +&91.50 $\pm$ 0.00 +&~7.53 $\pm$ 0.00 +\\
HOG&65.33 $\pm$ 0.00 +&69.33 $\pm$ 0.00 +&34.97 $\pm$ 0.00 +&15.07 $\pm$ 0.00 +&86.00 $\pm$ 0.00 +&19.67 $\pm$ 0.00 +\\
LBP&64.00 $\pm$ 0.00 +&54.00 $\pm$ 0.00 +&25.87 $\pm$ 0.00 +&15.28 $\pm$ 0.00 +&90.00 $\pm$ 0.00 +&39.33 $\pm$ 0.00 +\\
SIFT&86.67 $\pm$ 0.00 +&86.67 $\pm$ 0.00 +&55.94 $\pm$ 0.00 +&23.49 $\pm$ 0.00 +&98.50 $\pm$ 0.00 +&67.95 $\pm$ 0.00 +\\
LeNet&\textbf{91.78 $\pm$ 1.71} {--}&\textbf{88.42 $\pm$ 1.49} =&62.26 $\pm$ 3.67 =&40.85 $\pm$ 4.20 +&82.22 $\pm$ 2.78 +&55.37 $\pm$ 3.41 +\\
CNN&77.64 $\pm$ 9.59 +&79.49 $\pm$ 11.56 +&50.07 $\pm$ 4.27 +&32.09 $\pm$ 2.47 +&92.43 $\pm$ 1.68 +&65.80 $\pm$ 1.83 +\\
InceptionV3&61.38 $\pm$ 4.83 +&68.20 $\pm$ 6.67 +&51.23 $\pm$ 7.88 +&20.69 $\pm$ 4.59 +&94.88 $\pm$ 2.09 +&66.33 $\pm$ 8.78 +\\
InceptionResNetV2&50.00 $\pm$ 0.00 +&50.00 $\pm$ 0.00 +&19.14 $\pm$ 6.61 +& 15.31 $\pm$ 2.44 +&12.28 $\pm$ 20.88 +&~9.86 $\pm$ 13.01 +\\
\cellcolor[gray]{0.9}\textbf{KSMTGP}&\cellcolor[gray]{0.9}\textbf{87.91 $\pm$ 3.39}&\cellcolor[gray]{0.9}\textbf{88.24 $\pm$ 3.46 }&\cellcolor[gray]{0.9}\textbf{64.13 $\pm$ 2.60 }&\cellcolor[gray]{0.9}\textbf{46.45 $\pm$ 1.66 }&\cellcolor[gray]{0.9}\textbf{99.27 $\pm$ 0.53 }&\cellcolor[gray]{0.9}\textbf{99.35 $\pm$ 0.30}\\ \hline
Overall&13+, 1{--}&13+, 1=&12+, 1=, 1{--}&14+&14+&14+\\ \hline
\hline
\cellcolor[gray]{0.9}&\cellcolor[gray]{0.9}\textbf{KTH}&\cellcolor[gray]{0.9}\textbf{Outex}&\cellcolor[gray]{0.9}\textbf{Office\_D}&\cellcolor[gray]{0.9}\textbf{Office\_W}&\cellcolor[gray]{0.9}\textbf{COIL\_1}&\cellcolor[gray]{0.9}\textbf{COIL\_2}\\ \hline
SVM	&33.27 $\pm$ 2.53 +&23.78 $\pm$ 0.53 +&26.10 $\pm$ 0.52 +&34.10 $\pm$ 0.30 +&91.45 $\pm$ 0.00 +&96.15 $\pm$ 0.06 +\\
RF&53.93 $\pm$ 0.91 +&52.61 $\pm$ 0.36 +&42.83 $\pm$ 1.16 +&49.46 $\pm$ 0.67 +&93.97 $\pm$ 0.47 {--}&98.02 $\pm$ 0.36 +\\
KNN&28.69 $\pm$ 0.00 +&26.43
 $\pm$ 0.00 +&18.08 $\pm$ 0.00 +&23.59 $\pm$ 0.00 +&83.55 $\pm$ 0.00 +&84.52 $\pm$ 0.00 +\\
SRC&27.05 $\pm$ 0.00 +&~8.41 $\pm$ 0.00 +&21.57 $\pm$ 0.00 +&22.34 $\pm$ 0.00 +&90.00 $\pm$ 0.00 +&96.13 $\pm$ 0.00 +\\
LDA&30.00 $\pm$ 0.00 +&26.67 $\pm$ 0.00 +&23.62 $\pm$ 0.00 +&27.03 $\pm$ 0.00 +&87.90 $\pm$ 0.00 +&96.13 $\pm$ 0.00 +\\
DIF&50.98 $\pm$ 0.00 +&48.62 $\pm$ 0.00 +&29.74 $\pm$ 0.00 +&30.00 $\pm$ 0.00 +&84.52 $\pm$ 0.00 +&96.45 $\pm$ 0.00 +\\
Histogram&37.54 $\pm$ 0.00 +&71.04 $\pm$ 0.00 +&23.03 $\pm$ 0.00 +&23.12 $\pm$ 0.00 +&68.55 $\pm$ 0.00 +&83.39 $\pm$ 0.00 +\\
HOG&42.79 $\pm$ 0.00 +&21.82
 $\pm$ 0.00 +&30.61 $\pm$ 0.00 +&25.16 $\pm$ 0.00 +&62.92 $\pm$ 0.05 +&71.29 $\pm$ 0.00 +\\
LBP&74.26 $\pm$ 0.00 +&87.53
 $\pm$ 0.00 +&33.53 $\pm$ 0.00 +&37.97 $\pm$ 0.00 +&85.16 $\pm$ 0.00 +&97.58 $\pm$ 0.00 +\\
SIFT&74.75 $\pm$ 0.00 +&38.70 $\pm$ 0.00 +&48.98 $\pm$ 0.00 +&49.06 $\pm$ 0.00 +&92.74 $\pm$ 0.00 =&96.13 $\pm$ 0.00 +\\
LeNet&55.61 $\pm$ 4.75 +&72.86 $\pm$ 5.01 +&33.81 $\pm$ 2.53 +&36.15 $\pm$ 1.95 +&\textbf{96.02 $\pm$ 0.90} {--}&95.38 $\pm$ 0.95 +\\
CNN&41.28 $\pm$ 6.61 +&66.08 $\pm$ 7.99 +&40.85 $\pm$ 3.10 +&43.47 $\pm$ 1.99 +&92.77 $\pm$ 1.80 =&95.14 $\pm$ 1.09 +\\
InceptionV3	&55.50 $\pm$ 11.60 +&49.70 $\pm$ 16.33 +&48.04 $\pm$ 4.71 +&48.56 $\pm$ 2.54 +&90.82 $\pm$ 1.04 +&98.16 $\pm$ 0.66 +\\
InceptionResNetV2&25.65 $\pm$ 8.86 +&13.55 $\pm$ 10.83 +&~6.47 $\pm$ 6.94 +&~9.83 $\pm$ 9.62 +&36.95 $\pm$ 21.87 +&46.67 $\pm$ 22.83 +\\
\cellcolor[gray]{0.9}\textbf{KSMTGP}&\cellcolor[gray]{0.9}\textbf{94.14 $\pm$ 1.11}&\cellcolor[gray]{0.9}\textbf{98.76 $\pm$ 0.30 }&\cellcolor[gray]{0.9}\textbf{59.16 $\pm$ 2.10 }\cellcolor[gray]{0.9}&\cellcolor[gray]{0.9}\textbf{57.22 $\pm$ 1.74}&\cellcolor[gray]{0.9}\textbf{92.89 $\pm$ 0.81}&\cellcolor[gray]{0.9}\textbf{99.75 $\pm$ 0.41}\\ \hline
Overall&14+&14+&14+&14+&10+, 2=, 2{--}&14+\\
	\hline	\hline
		\end{tabular}	
		\label{table:testresultsOther1}
	\end{center}
\vspace{-6mm}
\end{table*}

\subsubsection{Comparisons with Multifactorial GP (MFFGP)} Table \ref{table:testresults} shows that KSMTGP achieves significantly better performance than MFFGP on eight datasets and similar performance on four datasets. On these 12 datasets, KSMTGP achieves better mean accuracy than MFFGP. KSMTGP also achieves better maximum accuracy than MFFGP on nine datasets. Comparing the results obtained by MFFGP with FGP, it can be found that the classification performance is slightly improved by simultaneously solving two tasks. However, the improvement is not significant. Compared with MFFGP, the proposed KSMTGP approach is more effective for multitask feature learning to image classification. Unlike an optimisation problem, which only needs to optimise the objective/fitness function, the feature learning and image classification problem need to improve the generalisation performance, e.g., the classification performance on the test set, which is vital and different from the objective/fitness function. Therefore, it is necessary to consider how to improve the generalisation performance when dealing with two learning tasks. The KSMTGP approach is able to learn both common and task-specific knowledge to describe features for image classification. The features described by the common tree learned from two tasks is more general, which potentially improves the generalisation performance.

\subsubsection{Comparisons with non-GP-based Methods} Since KSMTGP is proposed for image classification, it is necessary to compare it with existing image classification methods, including different classification methods (i.e. SVM, RF, KNN, SRC, and LDA), the methods using different manually extracted features (i.e. DIF, Histogram, HOG, LBP, and SIFT), CNNs (i.e. LeNet and CNN), and deep CNNs with transfer learning (i.e. InceptionV3 and InceptionResNetV2). This section further analyses the performance of KSMTGP by comparing it with these different comparison methods.

The classification results of KSMTGP and the 14 non-GP-based methods are listed in Table \ref{table:testresultsOther1}. KSMTGP achieves significantly better or similar performance in 164 comparisons out of the total 168 ($14\times 12$) comparisons. On several datasets, such as EYALE, KTH, Outex, Office\_D, Office\_W, the classification performance of these comparison methods are very low. The KSMTGP approach achieves much higher classification performance than these comparison methods, e.g., 8.69\% higher on EYALE, nearly 20\% higher on KTH, 11.23\% on Outex, 10.18\% on Office\_D, and 7.76\% on Office\_W. On a few datasets, some comparison methods such as LeNet and SVM achieve better performance than KSMTGP. However, the performance of these comparison methods varies with the datasets. For example, although LeNet achieves better performance than KSMTGP on three datasets (FEI\_1, FEI\_2, and COIL\_2), its performance on the other datasets (e.g., ORL, EYALE, KTH, Outex, Office\_D, Office\_W) is significantly lower than that of KSMTGP. Compared with these 14 comparison methods, KSMTGP has better adaptability since it is able to achieve better performance on these different types of image classification tasks. The comparisons and analysis indicate that KSMTGP is more effective and adaptive than these representative image classification methods on different types of image classification tasks. 

\begin{figure*}
	\centering
	\includegraphics[width=\linewidth]{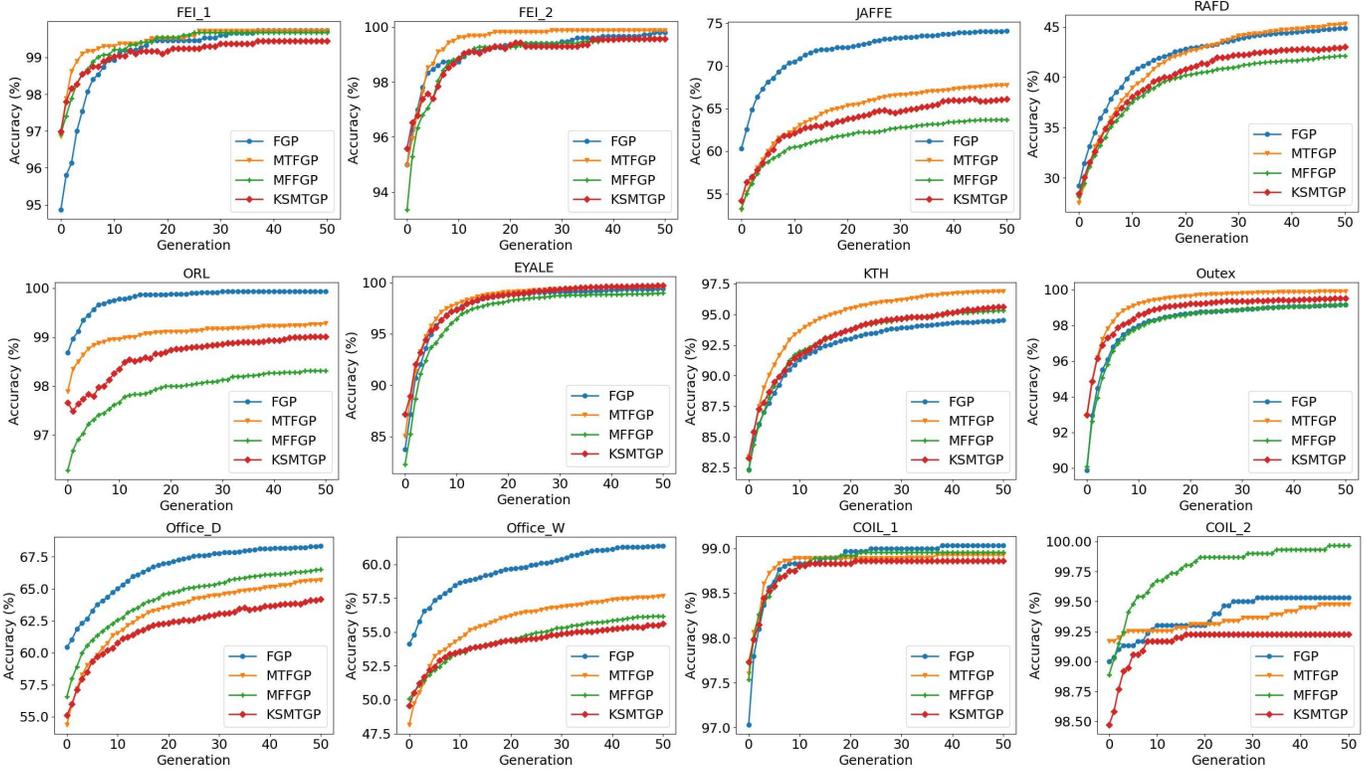}
	\vspace{-7mm}
	\caption{Convergence behaviours of the FGP, MTFGP, MFFGP, and KSMTGP methods over 50 generations on the 12 image classification datasets.}
	\label{fig:convergencecurve}
	\vspace{-6mm}
\end{figure*}
\vspace{-2mm}

\subsection{Training Results and Convergence Behaviours}
The average training results (i.e. fitness value according to Eq. (\ref{acc_tt1})) and the convergence behaviours of the GP-based methods, i.e., FGP, MTFGP, MFFGP, and KSMTGP, of the 30 runs are shown in Fig. \ref{fig:convergencecurve}. The single-tasking FGP method achieves the best fitness values than the other three methods on five datasets, i.e., JAFFE, ORL, Office\_D, Office\_W, and COIL\_1. The single-tasking FGP with a multi-tree representation achieves the best training results than the other three methods on four datasets, i.e., FEI\_2, RAFD, KTH and Outex. On the COIL\_2 dataset, the MFFGP method achieves the best training results. On the remaining two datasets, i.e., FEI\_1 and EYALE, the training results of these methods are very close. Compared with the three GP methods, the proposed KSMTGP approach achieves worse training results on almost all the datasets. However, KSMTGP achieves better testing (generalisation) performance than these three GP-based methods, as shown in Table \ref{table:testresults}. The results show that KSMTGP may be less overfitted to the training set by having a common feature set from the two tasks to perform image classification compared with these three methods. 

Fig. \ref{fig:convergencecurve} shows that the four GP-based methods have similar convergence behaviours, although they have different starting points due to the representation or searching mechanism differences. On most of these datasets, the four GP-based methods can converge to high fitness values, i.e., over 90\% accuracy. On some datasets, such as JAFFE, RAFD, Office\_D, and Office\_W, the single-tasking FGP method performs better than the other three GP methods. Different from these three methods, the best fitness values of the KSMTGP approach are not always increased over the generations. The best fitness values may decrease due to the change of the common tree in the individual representation, which leads to some fluctuations in the convergence curve. 

\subsection{Computational Time}
\begin{figure}[htbp]
		\vspace{-2mm}
	\centering
	\includegraphics[width=\linewidth]{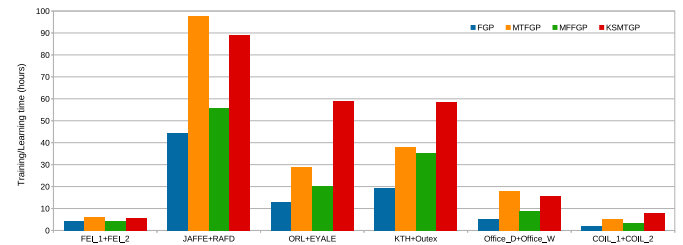}
	\vspace{-6mm}
	\caption{The average training/learning time (hour) of FGP, MTFGP, MFFGP, and KSMTGP on the six problems of 12 image classification datasets.}
	\label{fig:train_time}
	\vspace{-2mm}
\end{figure}

\begin{figure}[htbp]
	\centering
		\vspace{-4mm}
	\includegraphics[width=\linewidth]{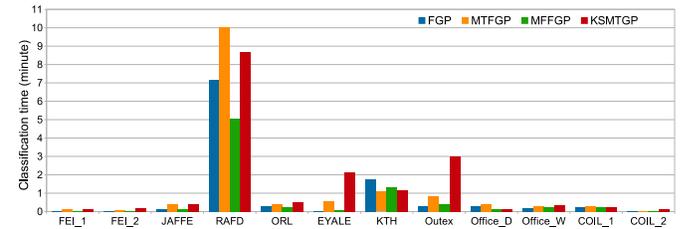}
	\vspace{-6mm}
	\caption{The average testing/classification time (minute) of FGP, MTFGP, MFFGP, and KSMTGP on the 12 image classification datasets.}
	\label{fig:test_time}
	\vspace{-2mm}
\end{figure}

The average training (i.e. evolutionary learning) time and the testing time (i.e. classification time) of the four GP-based methods are shown in Figures \ref{fig:train_time} and \ref{fig:test_time}. Note that the training time is the sum of the training time for the two datasets because MFFGP and KSMTGP solve two tasks simultaneously. Compared with FGP and MFFGP, which use one tree, the MTFGP and KSMTGP methods that use two trees as a solution to a task need longer training time on all the datasets. It is reasonable because adding one more tree will increase the time of fitness evaluation, including the time of feature transformation. Compared with MTFGP, KSMTGP uses longer training time on four problems (eight datasets) and less training time on two problems (four datasets). The reason may be that KSMTGP evolves some complex trees to achieve good performance. As it can be seen from the testing time in Fig. \ref{fig:test_time} that KSMTGP uses slightly longer time on classifying the EYALE and Outex datasets. However, since the training of these methods can be offline and the testing process is very fast so the training time is not so important. But the analysis still provides insights on the computational cost of these methods.

The testing/classification time is more important in real-world applications. Although KSMTGP uses longer training time, its overall testing time is short. Except for the RAFD dataset, KSMTGP uses less than four minutes on the other 11 datasets. It also can be found that the testing time of KSMTGP is close to that of the other three GP-based methods. RAFD is a large dataset with 7,240 testing images so that these GP-based methods use longer testing time. The comparisons show that KSMTGP is fast on most of the datasets in testing even with the use of two evolved trees to extract features for image classification.

\subsection{Number of Learned Features}
\begin{figure}[htbp]
	\vspace{-4mm}
	\centering
	\includegraphics[width=\linewidth]{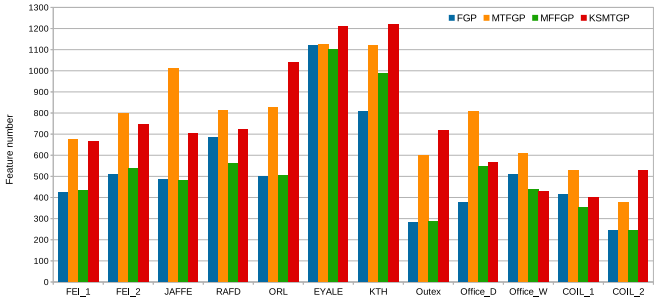}
	\vspace{-6mm}
	\caption{The average number of features learned by FGP, MTFGP, MFFGP, and KSMTGP on the twelve datasets.}
	\label{fig:feautrenumber}
	\vspace{-4mm}
\end{figure}

 The average number of features learned by the four GP-based methods are compared in Fig. \ref{fig:feautrenumber}. The four GP-based methods are able to learn a variable number of features from different image datasets. The feature number ranges from 200 to 1,200 on different datasets. Compared with FGP and MFFGP, MTFGP and KSMTGP learn a larger number of features for classification. It is reasonable because MTFGP and KSMTGP use two trees to extract features from an image, while FGP and MFFGP use one tree. By learning a larger number of features, MTFGP and KSMTGP achieve better generalisation performance than FGP and MFFGP, as shown in Table \ref{table:testresults}. However, a large number of features does not necessarily improve the generalisation performance. Compared with MTFGP, KSMTGP learns a smaller number of features on the FEI\_2, JAFFE, Office\_D, Office\_W, COIL\_1 datasets but achieves better generalisation performance. On the EYALE, KTH and COIL\_2 datasets, KSMTGP learns a larger number of features and achieves better performance than MTFGP. The results show that KSMTGP is able to learn a reasonable number of features to achieve better generalisation performance than the other three GP-based methods. 
 
\section{Further Analysis}
This section further analyses the trees, including the common and task-specific trees, evolved by the KSMTGP approach to show its effectiveness and generalisability. 

\subsection{Example Individual Analysis}

\begin{figure}[!htbp]
	\centering
	\vspace{-6mm}
	\includegraphics[width=0.9\linewidth]{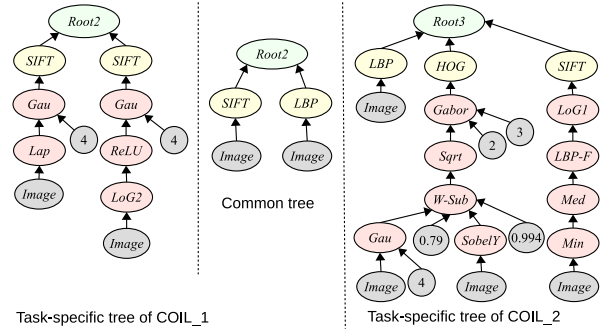}
	\vspace{-4mm}
	\caption{Visualisation of the three example trees on Problem 6.}
	\label{fig:problem6example}
	\vspace{-2mm}
\end{figure}

Two example individuals, i.e., three trees evolved by KSMTGP, on Problem 6, i.e., COIL\_1 and COIL\_2, are used for further analysis. The three trees are a common tree ($\Phi_{com}$) and two task-specific trees ($\Phi_{t1}$ and $\Phi_{t2}$). The $\Phi_{com}$ and $\Phi_{t1}$ trees are for solving the task of classifying the COIL\_1 dataset and achieve a classification accuracy of 93.06\%. The $\Phi_{com}$ and $\Phi_{t2}$ trees are for classifying the COIL\_2 dataset and achieve a classification accuracy of 100.0\%. These trees are listed as follows. The visualisation of them is shown in Fig. \ref{fig:problem6example}.  

\begin{multline}
\Phi_{t1}=Root2(SIFT(Gau(Lap(Image), 4)),\\ SIFT(Gau(ReLU(LoG2(Image)), 4)))
\end{multline}
\vspace{-4mm}
\begin{equation}
\Phi_{com}=Root2(SIFT(Image), HOG(Image))
\end{equation}
\vspace{-6mm}
\begin{multline}
\Phi_{t2}=Root3(LBP(Image), HOG(Gabor(Sqrt(W\text{-}Sub(\\Gau(Image, 1), 0.79, SobelY(Image), 0.994)), 2, 3)),\\ SIFT(LoG1(LBP\text{-}F(Med(Min(Image))))))
\end{multline}

The common tree $\Phi_{com}$ is simple and can extract a combination of SIFT and HOG features from an input image ($Image$). The number of extracted features are 128+64=192. From the results in Table \ref{table:testresults}, it can be found that the SIFT features perform better than the other features on the COIL\_1 and COIL\_2 datasets. In addition, the HOG features can capture the shape and appearance of the objects in COIl\_1 and COIL\_2 images. Therefore, the common tree has the $SIFT$ operators to extract a combination of SIFT and HOG features as the shared feature representation crossing the two tasks. 

The $\Phi_{t1}$ tree is slightly more complex than the common tree and can extract a combination of SIFT features from an input image. Before feature extraction, a number of filtering operations, including Laplacian filtering ($Lap$), Gaussian filtering ($Gau$) and Laplacian of Gaussian filtering ($LoG2$), are used to process the input image. The $\Phi_{t1}$ tree is able to produce 256 (128$\times$2) features from an input image. In total, 448 features are extracted for classifying the COIL\_1 dataset. The $\Phi_{t2}$ tree is for classifying the COIL\_2 dataset. It can extract a combination of LBP, HOG and SIFT features, i.e., 251 (59+64+128) features. The LBP features are extracted from the original images and the HOG and SIFT features are extracted from the images processed by different filtering operations, including Gaussian filtering, Gabor filtering ($Gabor$), median filtering, and Laplacian of Gaussian filtering. In total, 443 features are extracted for classifying the COIL\_2 dataset. 

The analysis shows that KSMTGP learns a simple but effective common tree crossing two image classification tasks, which is expected since the common tree is to extract features general to both tasks. KSMTGP evolves a complex task-specific tree together with the common tree to effectively solve a task. In addition, KSMTGP is able to evolve common and task-specific trees that produce a flexible number of features from images.

\subsection{Analysis on Common and Task-specific Trees}
The performance of the common and task-specific trees on the same dataset where they were learned from is analysed. Furthermore, the common and task-specific trees are transferred to solve other image classification tasks, which are different from the ones they learned from, to further show their generality and transferability.

\subsubsection{Classification Performance on the Same Datasets}
The classification performance of the common and task-specific tree is further analysed and compared to show the effectiveness of the new individual representation in the proposed KSMTGP approach. Table \ref{table:testresultsTrees} shows the classification results obtained by using these trees from the 30 runs on the dataset where they were learned from. Specifically, the classification accuracy is obtained by using one of these two trees to extract features for classifying the same datasets.

Table \ref{table:testresultsTrees} shows that the classification performance of the two trees (i.e., ``Overall") is better than that of the common tree and the task-specific tree on the majority of the datasets. Using two trees can represent more information (i.e., features) of the datasets, which potentially increase the classification performance. The current search mechanism in KSMTGP is able to produce the two trees that perform the best on the training set. As a result, using the combination of the features extracted by the two trees can achieve better performance than using those by each single of the two trees.

\begin{table}[htbp]
	\vspace{-4mm}
	\caption{Classification Accuracy (\%) obtained by two trees, the common tree, and the task-specific tree on each dataset}
	\vspace{-4mm}
	\begin{center}
		\begin{tabular}{|l|l|l|l|}
			\hline 
			&\textbf{Two trees}&\textbf{Common Tree}&\textbf{Task-specific Tree}\\
	&Mean $\pm$ St.dev&Mean $\pm$ St.dev&Mean $\pm$ St.dev\\\hline
FEI\_1&\textbf{87.91 $\pm$ 3.39} &84.36 $\pm$ 2.85& 84.07 $\pm$ 6.83 \\
FEI\_2&\textbf{88.24 $\pm$ 3.46}&86.78 $\pm$ 1.21&83.64 $\pm$ 7.22 \\
JAFFE&64.13 $\pm$ 2.60 &\textbf{64.29 $\pm$ 1.84}&54.41 $\pm$ 11.90\\
RAFD&\textbf{46.45 $\pm$ 1.66}&35.06 $\pm$ 1.42&42.44 $\pm$ 3.03\\
ORL&\textbf{99.27 $\pm$ 0.53}&97.82 $\pm$ 0.81 &98.90 $\pm$ 1.21\\
EYALE&\textbf{99.35 $\pm$ 0.30}&95.26 $\pm$ 0.87&99.05 $\pm$ 0.60\\
KTH&\textbf{94.14 $\pm$ 1.11}&87.65 $\pm$ 2.70&91.02 $\pm$ 3.46\\
Outex&\textbf{98.76 $\pm$ 0.30}&93.46 $\pm$ 1.42 &96.69 $\pm$ 2.48\\
Office\_D&\textbf{59.16 $\pm$ 2.10}&56.84 $\pm$ 0.99&46.99 $\pm$ 10.52\\
Office\_W&\textbf{57.22 $\pm$ 1.74}&55.26 $\pm$ 0.48&44.35 $\pm$ 8.31\\
COIL\_1&\textbf{92.89 $\pm$ 0.80}& 92.58 $\pm$ 0.00&86.88 $\pm$ 10.92\\
COIL\_2&\textbf{99.75 $\pm$ 0.41}&97.42 $\pm$ 0.00&97.81 $\pm$ 2.00 \\ \hline 
		\end{tabular}	
		\label{table:testresultsTrees}
	\end{center}
	\vspace{-4mm}
\end{table}

Table \ref{table:testresultsTrees} shows that the classification performance of the task-specific trees is better than that of the common trees on six datasets, i.e., JAFFE, RAFD, ORL, EYALE, KTH, Outex, and COIL\_2. The performance of the common trees and the task-specific trees varies with datasets. An important finding is that the performance of the common trees is more stable than that of the task-specific trees on each dataset because the standard deviation values obtained by the common trees are smaller than those by the task-specific trees. During the evolutionary process, the performance of the common trees is evaluated on the two tasks (in Algorithm \ref{FitnessEvaluationCommon}), while the performance of the task-specific trees is jointly evaluated with the common tree (in Algorithm \ref{FitnessEvaluation}). This leads to the common trees themselves are more effective and general than the task-specific trees. 

\subsubsection{Classification Performance on the Other Datasets}
The common and the task-specific trees learned from one dataset are applied to extract features for classifying another different dataset to further analyse their generality and transferability. Table \ref{testresultsOtherdatasetsTranfer} lists the results obtained by these trees transferred from one dataset to another dataset. The results are from the trees of the 30 independent runs. The experiments are conducted crossing the same type of image classification tasks, i.e., facial expression classification and object classification. In the first column of Table \ref{testresultsOtherdatasetsTranfer}, ``X$\rightarrow$Y" indicates that the trees learned from the X dataset are used to classifying the Y dataset. Note that this section aims to further analyse the performance of the trees learned by KSMTGP to provide more insights into the performance and behaviours of KSMTGP.

\begin{table}[htbp]
	\vspace{-4mm}
	\caption{Classification accuracy (\%) obtained by applying the GP trees learned from one dataset to another different dataset}
	\vspace{-6mm}
	\setlength\tabcolsep{2.4pt} 
	\begin{center}
		\begin{tabular}{|l|l|l|l|}
			\hline 
			Datasets&\textbf{Two trees}&\textbf{Common Tree}&\textbf{Task-specific Tree}\\
			&Mean $\pm$ St.dev&Mean $\pm$ St.dev&Mean $\pm$ St.dev\\\hline
			FEI\_1$\rightarrow$ JAFFE&55.66 $\pm$ 3.36&57.78 $\pm$ 1.63&50.44 $\pm$ 7.36\\
			JAFFE$\rightarrow$ FEI\_1&87.64 $\pm$ 3.67&87.05 $\pm$ 2.75&81.35 $\pm$ 8.42\\
			FEI\_2$\rightarrow$ RAFD&27.50 $\pm$ 3.89 &25.28 $\pm$ 1.62&23.89 $\pm$ 4.59\\
			RAFD$\rightarrow$ FEI\_2&85.69 $\pm$ 2.13 &84.76 $\pm$ 2.64&84.22 $\pm$ 2.40\\
			Office\_D$\rightarrow$ COIL\_1&91.67 $\pm$ 1.08&92.15 $\pm$ 0.32&89.02 $\pm$ 3.19\\
			COIL\_1$\rightarrow$ Office\_D &49.72 $\pm$ 3.28&48.69 $\pm$ 0.00&37.68 $\pm$ 11.10\\
			Office\_W$\rightarrow$ COIL\_2&99.83 $\pm$ 0.35&99.86 $\pm$ 0.44&96.33 $\pm$ 9.59\\
			COIL\_2$\rightarrow$ Office\_W&52.88 $\pm$ 5.13&50.62 $\pm$ 0.00&39.62 $\pm$ 8.61\\
			\hline
		\end{tabular}	
		\label{testresultsOtherdatasetsTranfer}
	\end{center}
	\vspace{-4mm}
\end{table}

Comparing the results in Table \ref{testresultsOtherdatasetsTranfer} with those in Table \ref{table:testresultsOther1}, it can be found that these transferred GP trees can achieve better performance than most of the comparison methods, such as SRC, RF, KNN, LDA, Histogram, HOG, LBP, SIFT, InceptionV3, and InceptionResNetV2, on these datasets. Specifically, InceptionV3 and InceptionResNetV2 used the models pretrained on ImageNet to extract features from images for classification. Compared with InceptionV3 and InceptionResNetV2, the trees (i.e., feature extractors) learned by KSMTGP are more effective for solving other tasks by achieving better classification performance on the datasets different from the datasets being used to learn them. This indicates that the trees learned by KSMTGP have good transferability. 

The results in Table \ref{testresultsOtherdatasetsTranfer} show that the common trees achieve better classification performance than the task-specific trees on most datasets, including the FEI\_1, FEI\_2, JAFFE, EYALE, KTH, Office\_D, and COIL\_2 datasets. The common trees are more effective than the task-specific trees when applying them to solve other image classification tasks. This is because the common trees are jointly learned from the two tasks so that they can represent general features that are effective for multiple image classification tasks. The task-specific trees represent features that are specifically effective for a particular dataset. However, with image-related operators and descriptors as internal nodes, the task-specific trees still can represent effective features for image classification, although they are not as effective as the common trees. 

To sum up, the analysis shows that using both the common and task-specific trees can achieve better performance than using one of them, individually. Two trees can represent richer information of the datasets to achieve better generalisation performance. The common trees are learned jointly from the two tasks and can represent general but effective features crossing different tasks. Further analysis on the performance obtained by these trees on different datasets shows the high transferability of both the common and task-specific trees, and the common trees have better transferability than the task-specific trees due to the more general information learned from both tasks.

\section{Conclusions}
The goal of this paper was to develop a multitask GP approach to feature learning for image classification with limited training data. This goal has been successfully achieved by developing the KSMTGP method with explicit knowledge sharing. A new individual representation, a new evolutionary process, and new fitness functions were developed to allow KSMTGP to automatically evolve common trees and task-specific trees that can extract informative and effective features for image classification. The performance of KSMTGP was examined on six problems of 12 image classification datasets with limited training data and compared with a large number of competitive methods. 

The proposed KSMTGP method achieved better or similar generalisation performance than the single-tasking GP method, the single-tasking GP with a multi-tree representation method, the multifactorial GP method, and 14 non-GP-based methods. The results showed that the new individual representation with a common tree and the the knowledge sharing mechanism improve the generalisation performance of KSMTGP. Further analysis on example trees showed that KSMTGP learned a simple yet effective common tree for sharing crossing two tasks. Compared with the common tree. Compared with the task-specific trees, the common trees achieve better and more stable classification performance. In addition, the common trees have better transferability than the task-specific trees.

As a starting point, this study shows the potential ability of multitask GP for image feature learning. Unlike most existing evolutionary multitask methods, the proposed method can achieve explicit knowledge sharing by making the best use of the flexible representation of GP. Furthermore, the good transferability of GP tree has been demonstrated in this study. There is still a large research space in this direction. First, this paper focuses on simultaneously solving two tasks. It is also necessary to investigate whether new multitask GP can solve more than two tasks, simultaneously. Second, this paper treats a multitask problem as the same type of image classification tasks and allows GP to determine what to learn for sharing. It is also important to measure the task relatedness to set a multitask problem. However, designing such a measurement is very challenging because the image data are raw pixels and their feature space keeps changing during the feature learning process. In the future, we will investigate these potential directions.

\bibliographystyle{IEEEtran}
\bibliography{IEEEabrv,draft14.bib}
\end{document}